# A multiset model of multi-species evolution to solve big deceptive problems

**Luís Correia**[1], **António Manso**[1,2]

1- LabMAg - Laboratório de Modelação de Agentes, Faculdade de Ciências da Universidade de Lisboa, Campo Grande, 1749-016 Lisboa – Portugal

2 - Instituto Politécnico de Tomar, Quinta do Contador - Estrada da Serra, 2300-313 Tomar – Portugal

**Abstract** This chapter presents SMuGA, an integration of symbiogenesis with the Multiset Genetic Algorithm (MuGA). The symbiogenetic approach used here is based on the host-parasite model with the novelty of varying the length of parasites along the evolutionary process. Additionally, it models collaborations between multiple parasites and a single host. To improve efficiency, we introduced proxy evaluation of parasites, which saves fitness function calls and exponentially reduces the symbiotic collaborations produced. Another novel feature consists of breaking the evolutionary cycle into two phases: a symbiotic phase and a phase of independent evolution of both hosts and parasites. SMuGA was tested in optimization of a variety of deceptive functions, with results one order of magnitude better than state of the art symbiotic algorithms. This allowed to optimize deceptive problems with large sizes, and showed a linear scaling in the number of iterations to attain the optimum.

**Keywords:** Genetic Algorithm; Multisets; Symbiogenesis; Deceptive optimization problems

## Introduction

Computational models of coevolution can be used to study both natural settings and artificial scenarios. Moreover, they can solve optimization problems. Computational models are an effective tool configurable to model different types of multi-species evolution: parasitism, commensalism, mutualism and cooperative interactions. Competitive multi-species evolution has been useful in optimization applications due to it providing better results when compared with a single problem solver population. Coevolution of a solver population with a problem creator population pushes both populations to increasingly better solutions, a phenomenon called arms-race (Rosin and Belew, 1997).

On the other hand, symbiosis is a form of cooperative coevolution, which has been gaining relevance in biology (Daida et al., 1996). In artificial systems, symbiogenetic coevolution has been shown to improve evolutionary optimization algorithms by a specialization of the different components of the symbiotic collaboration (Wallin et al., 2005). In this case of cooperative coevolution there is a kind of division of labor between the different types of symbionts. Each host is combined with a set of parasites forming a collaboration. Each collaboration is evaluated as a solution to the optimization problem. This is repeated for different hosts and parasites. Artificial symbiogenetic evolution is proving useful in solving deceptive problems (Wallin et al., 2005), a class of functions that is especially difficult to optimize due to the fact that the optimum is surrounded by regions of low quality solutions.

Artificial evolutionary models are inspired by nature, but when used as engineering tools they do not need to maintain a strict correspondence with their natural counterparts. The main goal of engineering is to obtain efficient tools, in this case designed to solve optimization problems. Taking this into account, we further explore different approaches of evolutionary algorithms (EA) and their operators that one may consider unrealistic by comparison to nature. The multiset representation of populations is one of those examples and in previous work we have used that representation to support the evolutionary algorithms populations (Manso and Correia, 2009). That algorithm is called Multiset Genetic Algorithm (MuGA) and is successful in the optimization of various kinds of problems. The populations are represented by multisets and the operators that are used explore the representation in order to make the evolutionary process more efficient and effective in a optimization of difficult problems.

In this work we present the Symbiogenetic MuGA Algorithm (SMuGA), which uses natural inspiration of symbiogenesis to solve large deceptive problems that are not solved by the common version of MuGA.

In the next section we present the base algorithm MuGA. In the following section the symbiogenetic approach used is detailed. In particular we have two different evolutionary processes, one for the hosts and another for the parasites, and we describe each one separately and then aggregated. Next, we present results obtained in several types of deceptive functions. The final section of this chapter presents conclusion and proposes future work.

## MuGA - A Multiset Genetic Algorithm

MuGA is a genetic algorithm that explores the features of a multiset to represent populations of evolutionary algorithms and to improve their performance. The traditional representation of populations used in evolutionary algorithms raises two types of problems: the loss of genetic diversity during the evolutionary process and evaluation of redundant individuals. These problems can be alleviated when using multisets to represent populations.

Multiset population is not a representation that can be found in the natural world, but it works well for optimization of difficult engineering problems.

## *Populations represented be multisets*

A multiset (or multiple memberships set) is a collection of objects, called elements, which are allowed to repeat. We can define the multiset as a set of ordered pairs <*copies, element*> where copies are the cardinality associated to the element. MuGA is a genetic algorithm in which populations represented by multisets are called MultiPopulations (MP) and individuals represented by pairs <*copies, genotype*> are called MultiIndividuals (MI).

**Fig. 1** shows a simple population (SP) with eight individuals of OnesMax problem and the equivalent MP with four MI. A multiset representation of populations contains characteristics that make it a good alternative to the collections that are usually used:

– MI has always different genotypes and the size of MP corresponds to the genotype diversity at the genotypic level;
– The number of copies of MI may be used to control the selection pressure in favor of the best fit individuals;
– The compact representation needs less computational effort to store the population and avoids evaluation of identical individuals.

| Individual | Fitness |
|---|---|
| 11111110 | 7 |
| 11111110 | 7 |
| 11111110 | 7 |
| 11110000 | 4 |
| 11110000 | 4 |
| 10001001 | 3 |
| 10001001 | 3 |
| 10000000 | 1 |

a)

| MultiIndividual | Fitness |
|---|---|
| < 3, 11111110 > | 7 |
| < 2, 11110000 > | 4 |
| < 2, 10001000 > | 3 |
| < 1, 10000000 > | 1 |

b)

**Fig. 1.** a) Simple Population of 8 individuals; b) MultiPopulation of 4 MultiIndividuals.

The introduction of individuals in a MP is done either by incrementing the number of copies of corresponding MI if the genotype exists in the population or by introducing a new pair <*1,genotype*>. The elimination is done by decrementing the number of copies of corresponding MI if the number of copies is greater than one, or otherwise by removing the MI.

## *MuGA Algorithm*

In evolutionary algorithms, populations are traditionally represented as a collection of individuals. To minimize the issues such models raise, we developed MuGA (Algorithm 1), whose most distinctive feature is that it represents populations by multisets.

The algorithm starts by randomly generating and evaluating ***n*** individuals of the ***problem*** to be optimized, while assuring that the base population, MP0, contains ***n*** different genotypes. The design of the MuGA algorithm is prepared to preserve the genetic diversity by maintaining the dimension of MP0 across generations. The evolutionary process starts by selecting ***m*** individuals from MP0. These ***m*** individuals are stored in MP1 and the number of MI is less than or equal to ***m***. The process continues with the recombination of MP1 and subsequent mutation of MP2, generating MP3. MP4 is produced by the application of the replacement operator on MP0 and MP3 to select ***n*** MI from the two populations. This operator maintains the number of MI as a constant across generations. The evolutionary process tends to produce many copies of good individuals. To reduce the number of copies in MP4 the rescaling operator is applied and produces a new population (MP0) for the iterative evolutionary process.

```
MuGA (n , m , problem)
  MP0 = generate n MultiIndividuals from problem
  Evaluate MP0
  Repeat
     MP1 = Select m Individuals from MP0
     MP2 = Recombine the Individuals of MP1
     MP3 = Mutate the Individuals of MP2
     Evaluate MP3
     MP4 = Select n MultiIndividuals from MP3 and MP0
     MP0 = Rescale the number of copies of MP4
  Until stop criteria
End Function.
```

**Algorithm 1**- MuGA - Multiset Genetic Algorithm

Multipopulations enable the execution of traditional genetic operators and allow the design of new operators using the extra information, a set of unique genotypes and associated number of copies, to extend operators that benefit from such information. Next, we briefly describe the behavior of genetic operators using MPs.

## *Multiset Selection*

This operator chooses, from the base population, the parents that will be reproduced to generate new individuals. We first expand the MP to an SP, **Fig. 1**, so that MI with multiple copies has higher probability of being selected. We can then use traditional selection operators (tournament selection, proportional selection or ranking selection) or any improved selection operator (Sivaraj and Ravichandran, 2011). When the operator allows the selection of the same individual several times over, the mating population will contain MI and the number of copies will reflect the degree of fitness of the genotype. The number of copies of the fittest individuals tends to be larger than the remaining elements and can be explored by the subsequent genetic operators.

## *Multiset Recombination*

The recombination operator is responsible for the combination of chromosomes to produce offspring that share genetic material of both parents. There is a great variety of recombination operators in accordance with the representation of the genes and chromosomes (e.g. binary strings, vectors of real numbers or trees) of individuals and the type of problem to be solved, e.g. optimization of real functions (Herrera et al., 2003), permutations (Otman and Jaafar, 2011) or combinatorial (Spears and Anand, 1991). All these operators can be used in MuGA through equivalence between MP and SP in terms of genotype representation. Nevertheless we can design new operators using the number of copies to make a genotype associated with the various parameters of the genetic algorithm such as the probability of application, the number of cutting points, the strength of individuals to spread their genes, etc. A wide range of possibilities is available to explore the usefulness of this information and (Manso and Correia, 2011) presents a multiset recombination operator applied to the optimization of real coded functions.

## *Multiset Mutation*

The mutation operator in EA mimics what occurs in nature and randomly changes a (usually small) part of the genome. The main function of this operator is the introduction of new genes, enabling exploration of new areas in the search space that are not attainable by the recombination of parental characteristics. Like the recombination operator, mutation is also dependent on the type of problem and representation of the individuals (Abdoun et al., 2012), (Droste et al., 2002). A new operator that uses multiset information to optimize deceptive binary functions, called Multiset Wave Mutation (MWM) is presented in (Manso and Correia, 2013) and another one used to optimize real coded functions is presented in (Manso and Correia, 2011).

### *Multiset Replacement*

After recombination and mutation, the evolutionary algorithm has two populations of individuals: the main population and the offspring generated by genetic operators. The replacement operator selects which individuals will continue in the evolutionary process. The generational strategy replaces the parents with their children and the steady-state strategy replaces only a few parents with offspring (Lozano et al., 2008). The operator must maintain the genetic diversity in the main population so that the genetic operators can circumvent local optima and avoid premature convergence (Yu and Suganthan, 2010), (Jayachandran and Corns, 2010). A new operator that uses multiset information to replace populations in a steady state strategy, called Multiset Decimation Replacement (MDR), is presented in (Manso and Correia, 2013).

### *Multiset Rescaling*

The introduction of repeated elements in the MP tends to increase the number of copies of the best fit MI if nothing is done to oppose it.

The rescaling operator was proposed to avoid that the best individuals get too many copies (Manso and Correia, 2009). In order to control the number of repeated elements, the rescaling operator divides the number of copies of each MI by a factor, controlling in this way the pressure exhibited by the fittest individuals. The operator ensures that each MI has at least one copy and that the total number of individuals in the MP is not greater than a constant. An adaptive form of this operator, called Adaptive Rescaling (AR) calculates in each iteration the value of the reduction factor to maintain approximately the desired number of individuals.

## SMuGA – A Symbiogenetic Multiset Genetic Algorithm

Symbiosis is set of natural theories that try to explain the natural relationship between individuals that live together and how that relationship is vital to the survival of the group. In nature symbiosis occurs and involves a relationship that is constant and intimate between dissimilar species (Daida et al., 1996). That relationship is more than the ecological interaction and includes mutualism, where both individuals gain advantages from the alliance; commensalism, in which one individual gains advantages and the other doesn't have any inconvenience; and parasitism, where one individual gains advantages and the other is harmed by the relation.
Symbiosis theory provides an additional genetic operator to the artificial evolutionary process and is successfully applied to solve a wide range of hard problems. See (Heywood and Lichodzijewski, 2010) for a review of symbiogenesis as a mechanism to build complex adaptive systems.

The Symbiogenetic MuGA algorithm (SMuGA) is inspired by the Symbiogenetic Coevolutionary Algorithm (SCA), proposed by Wallin et al. in 2005, which explores a host-parasite relationship for optimization of concatenated deceptive functions. Although the names "hosts" and "parasites" suggest a parasitic relationship, the interaction between two species is benign and the gains of parasites are not harmful to the hosts. SCA is successfully used to optimize concatenated deceptive functions and MuGA by itself has proved to be an efficient algorithm in the optimization of such functions with a moderate size (Manso and Correia, 2013).

However, when the size of the problems increases MuGA experiences difficulties in its optimization. In this paper, we apply the concept of symbiosis to increase the efficiency of the MuGA. SMuGA is an algorithm that uses two cooperative species, hosts and parasites, which evolve together in a mutualistic relationship. The parasites are composed of a tuple <*position*, *genome*>, where the position represents the parasite genome location where the parasite acts, and the genome represents the genetic material of the parasite. In SMuGA the host genome is replaced by the genome of the parasite in the location defined by the position attribute (**Fig. 2**). The parasite considers the host genome as a circle, which means that when the copy of the parasite genome to the host reaches the limit of the host genome, the copy continues in the beginning. In **Fig. 2**, parasite ***p1*** is applied in host genome alleles 4, 5 and 6 and parasite ***p2*** is applied in the host genome alleles 9 and 0. The collaboration is the combination of host genes and the genes introduced by parasites p1 and p2.

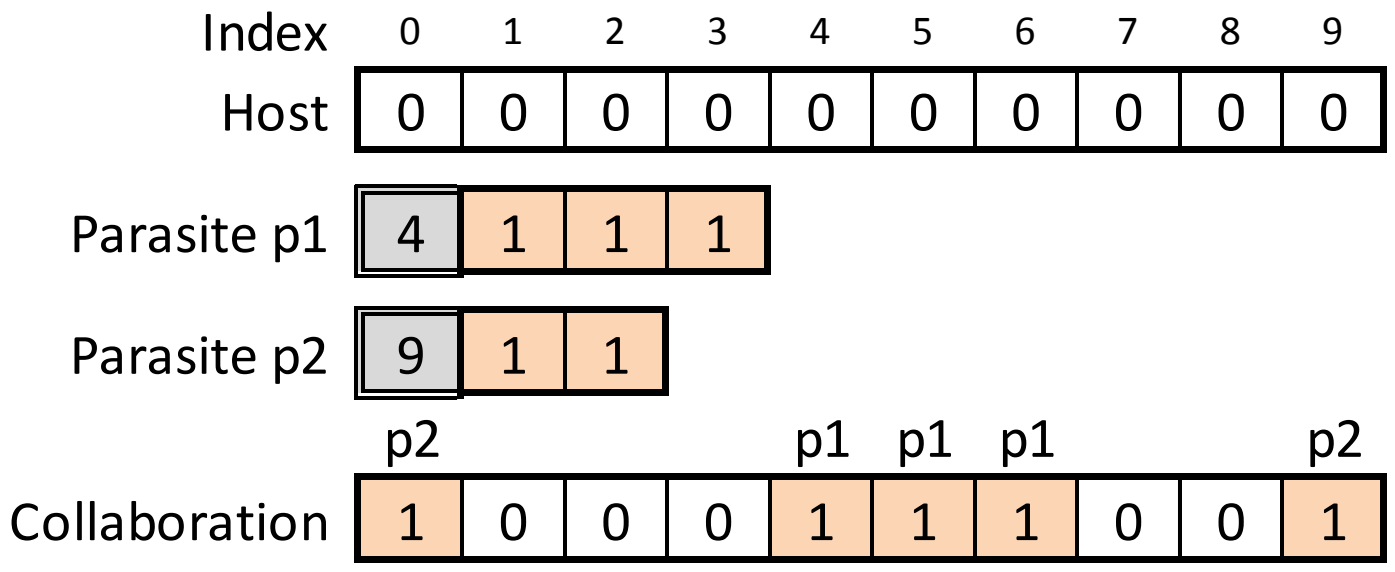

**Fig. 2.** Collaboration formed by the symbiosis of a host and a parasite.

SCA has some deficiencies identified by the authors. The size of the parasites is static and defined as a parameter, and collaboration is from one parasite to one host, where each host can only be infected by a parasite at a time. The best results obtained by the algorithm are when the parasite genome size is similar to the size of the functions to be optimized, the building blocks (BB), and the performance degrades quickly as the size of the parasites deviates from the size of the BB. Another weakness of the SCA algorithm is that the collaboration is one to one, which limits its applicability to separable problems.

The SMuGA was designed to suppress these two shortcomings by combining the concept of symbiosis with the potential that the populations based on multisets present on the optimization of this kind of functions. In the next section we present the representation and evolution of parasite populations, the evolution of host populations and the interaction between them with Symbiogenetic Multiset Genetic Algoritm (SMuGA). In the design of the SMuGA some choices are made with the objective of enhancing the success of the algorithm in the optimization of problems and contouring the shortcomings that SCA presents.

## *Evolution of Parasites*

In order to avoid having a human choice interfere significantly in performance, we eliminate the need to specify the size of the parasites. As mentioned earlier the work of Wallin et al (2005) showed that there was a very strong dependence of performance relative to the size of the parasite. When the size of parasites approaches BB size the performance is good, however it decays very quickly with deviations from the ideal dimension.

In our approach the user does not have to know the size of BB because the algorithm adapts the parasite's length as necessary. To the best of our knowledge, this is the first model of parasites that may vary their length along the evolutionary process. This system is important for solving problems in which the size of BB is not known or the BB has a variable size. The size of the parasites is changed by genetic operators of recombination and mutation. The selection operator gives opportunity to parasites that have a good performance in the host population to reproduce and to pass on their genetic material and position to their descendants. According to the theory of survival of the fittest, the parasites with a good genome, which includes the position of application and the genetic material, will spread their genes to subsequent generations, discovering and optimizing simultaneously the position, the size and alleles of the parasites.

### Parasite recombination

The following four situations can occur when two parasites recombine:

1. The parasites do not share positions in the genome of the host;
2. The parasites occupy consecutive positions in the genome of the host;
3. The parasites share some positions in the host;
4. All positions of one of the parasites occupy positions of the other.

In the first case, as the parasites infect different regions of the host genome, recombination between the two parasites cannot take place. In all other cases the idea underlying this operator is not only to recombine genetic material but also to introduce different genome lengths. We selected the recombination of parasite genomes as the principal operator to grow and shrink the length of the parasites.

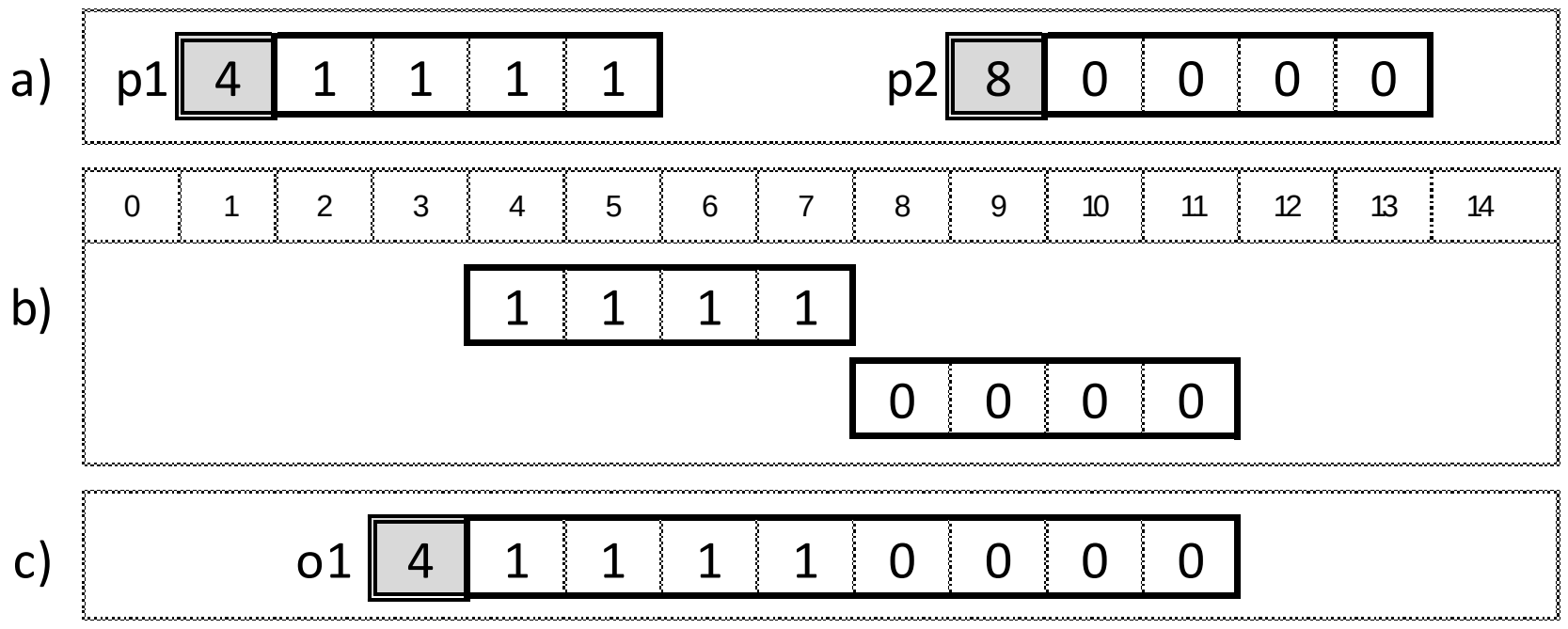

**Fig. 3.** Recombination by the union of consecutive parasites: a) Selected parasites; b) Positions occupied by parasites in the genome; c) Result of p1 and p2 recombination.

In the second case, **Fig. 3**, in which the parasites occupy consecutive locations in the host genome, we determine that recombinant parasites are the union of genomes generating a single parasite. The offspring ***o1***, **Fig. 3** c), has a genome whose size is the sum of the size of the parental genomes. This type of reproduction connects the parasites, and increases the length of the parasite genome.

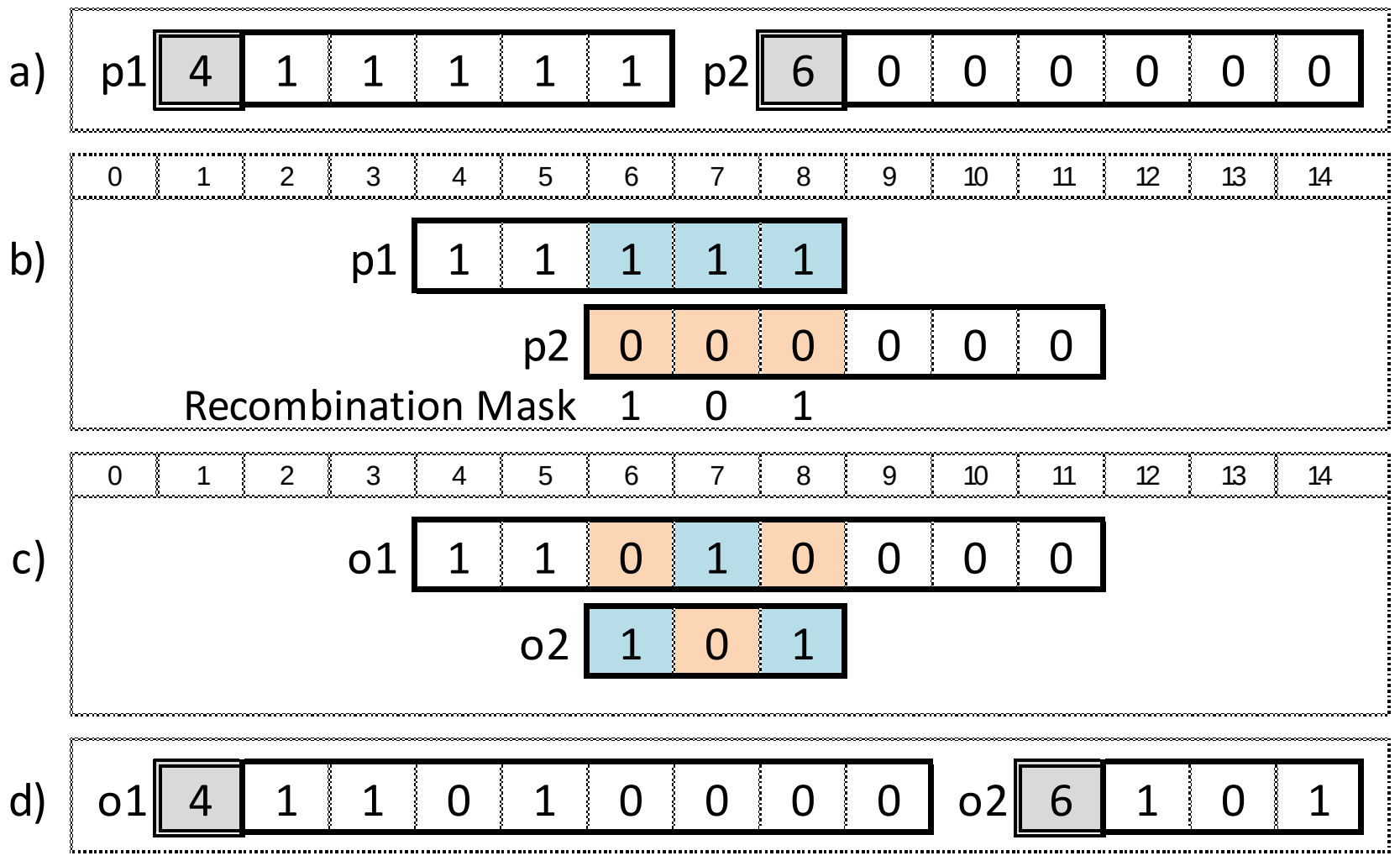

**Fig. 4.** Recombination by the share of some positions in the host: a) Selected parasites; b) Positions occupied by parasites in the genome; c) Result of p1 and p2 recombination; d) offspring parasites.

In case 3, **Fig. 4**, in which the parasites share some positions in the host genome, alleles in the overlapping zone are combined using uniform crossover. Furthermore, the offspring will have different genome sizes compared to their parents. In **Fig. 4** b) we illustrate uniform crossover. A recombination mask is randomly obtained to perform an exchange of the parental alleles in the overlapping zone. The symbol ***1*** in the mask means that there is an exchange of alleles in the overlapping zone and the symbol ***0*** means the opposite. **Fig. 4** c) and d) show the recombination result of parents ***p1*** and ***p2***. The offspring ***o1*** inherits from both parents the parts that are not common between them, as well as the recombined genome produced by the recombination mask. The offspring ***o2*** inherits only the recombined common part with a dual mask. The offspring ***o1*** is longer than the parents and ***o2*** is shorter.

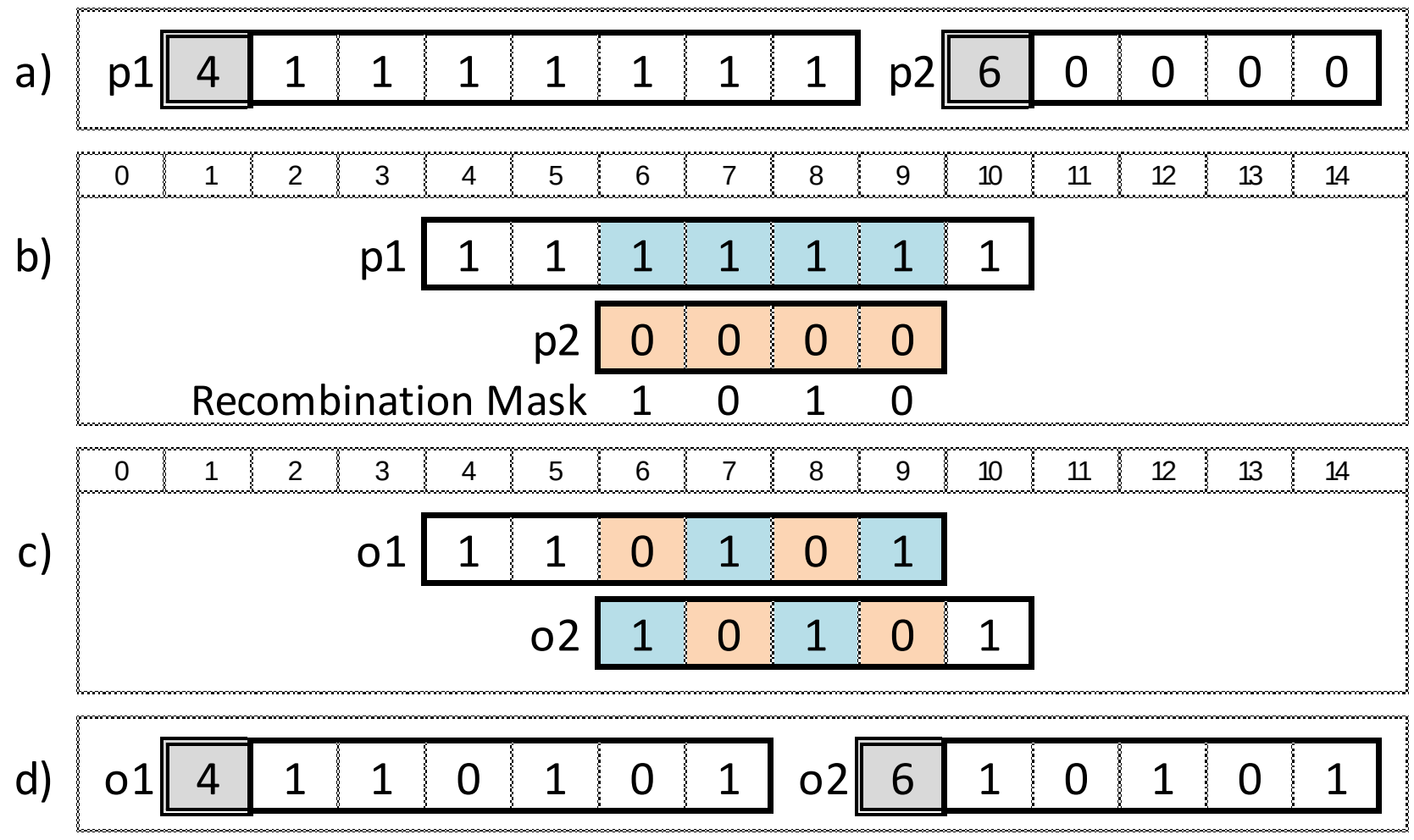

**Fig. 5.** Recombination when one of parasites occupies all the positions of other: a) Selected parasites; b) Positions occupied by parasites in the genome; c) Result of p1 and p2 recombination; d) offspring parasites.

In case 4, **Fig. 5**, where one of the parasites, ***p1***, occupies all the positions of the other, ***p2***, in the genome of the host, the overlapping zone is also recombined using uniform crossover. As in the previous case the genetic material is exchanged in the overlapping zone through a recombination mask, **Fig. 5** b) generated from a uniform distribution. **Fig. 5** d) shows the result of the recombination and the complete offspring. Individual ***o1*** inherits from the parent ***p1*** the first part not common to both parents, and the recombined common part, and the offspring ***o2*** inherits the dual recombined common part, and the last not common part of ***p1***. In this case the small parasites act as cutting knives of larger parasites.

## *Parasite Mutation*

The mutation operator randomly changes features of a parasite. These features include the position, length and their genetic material. We use three types of parasite mutation:

1. Change in anchoring position;
2. Change in the genome;
3. Parasite genome splitting whereby two new parasites are formed.

In the first situation, parasites change the position of host infection. In **Fig. 6** a) the parasite ***p1*** that infects the fourth position generates the mutant ***m1*** infecting position 10 with the same genotype. Note that the parasite ***m1*** affects the host genome in a circular way where the last three bits of the parasite infect the first three positions of the host.

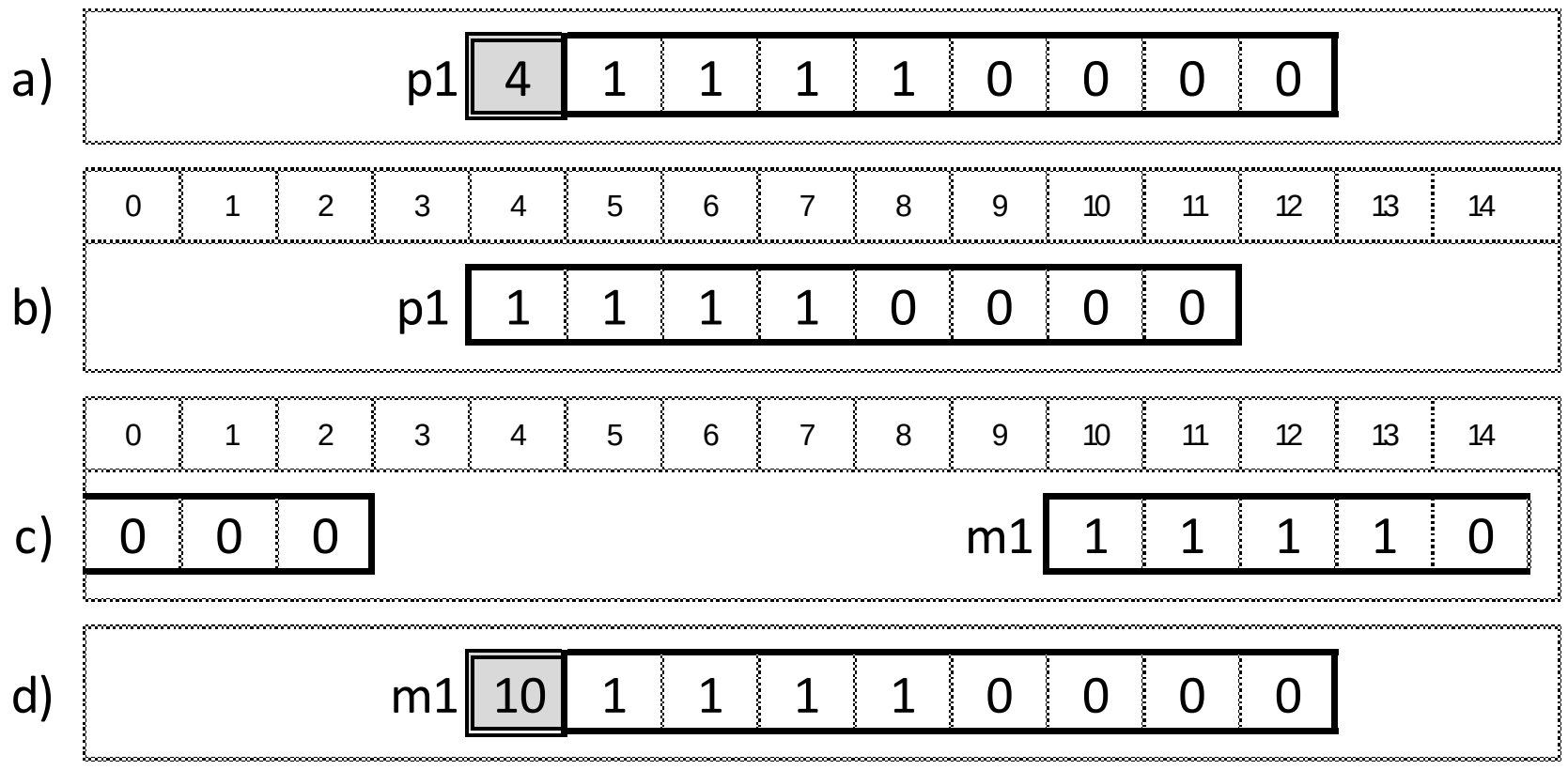

**Fig. 6.** Mutation by changing position: a) Original parasite; b) Positions occupied by original; c) Positions occupied by mutant parasite; d) Mutant parasite.

In the second case, the value of the alleles is changed by a probability distribution that generates the mutation mask shown in **Fig. 7** b. At the positions where the mask has the value ***1*** the bit value of parasite is flipped. In this situation, only the value of the parasite's genome is modified, which enhances the appearance of parasites in the population with new genomes.

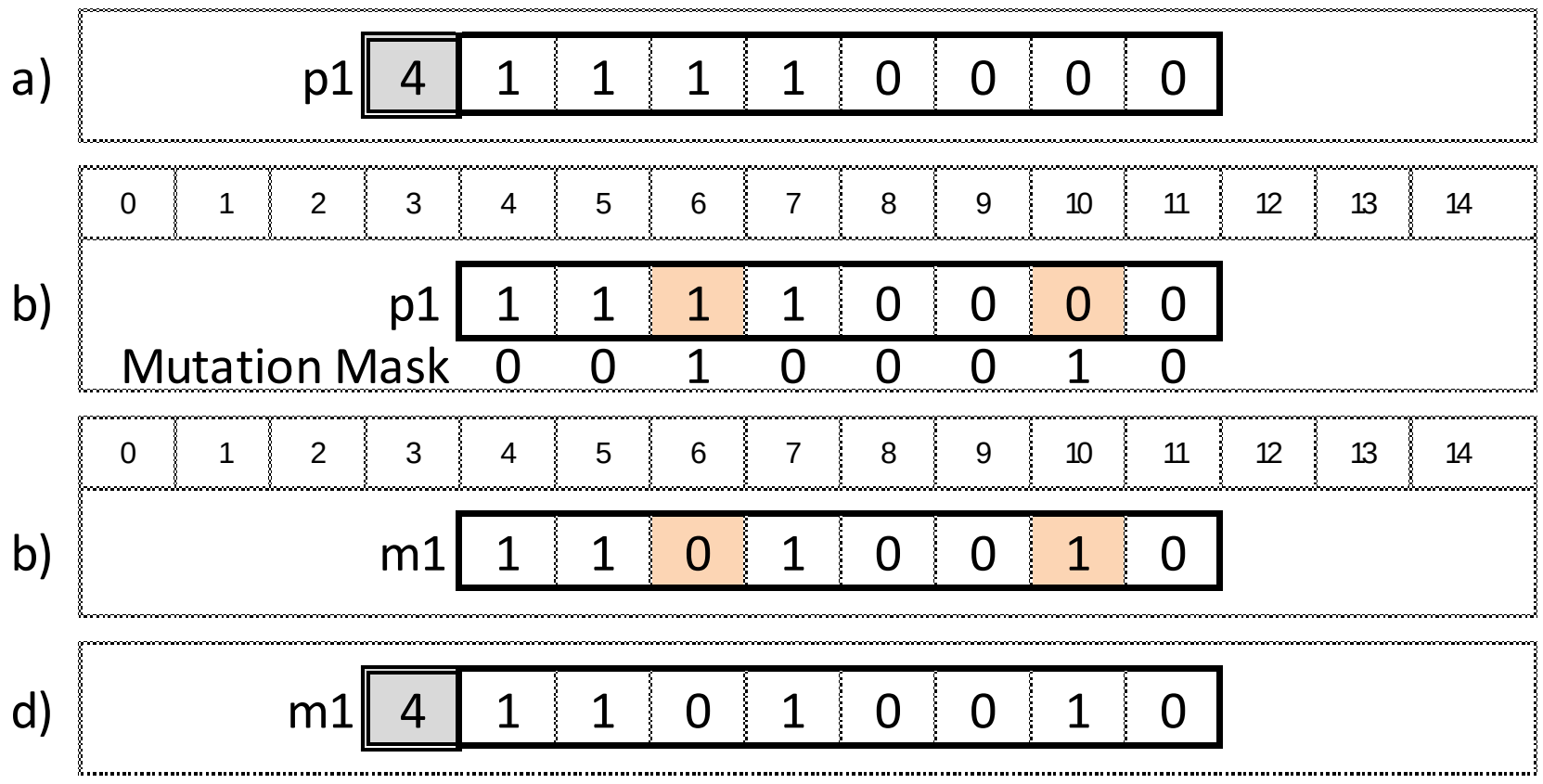

**Fig. 7.** Mutation by changing genome: a) Original parasite; b) Positions occupied by original and mutation mask; c) Positions occupied by mutant parasite; d) Mutant parasite.

In the latter situation the parasite genome is split into two parts, originating into two new parasites. The probability to split a genome is proportional to its length in bits.

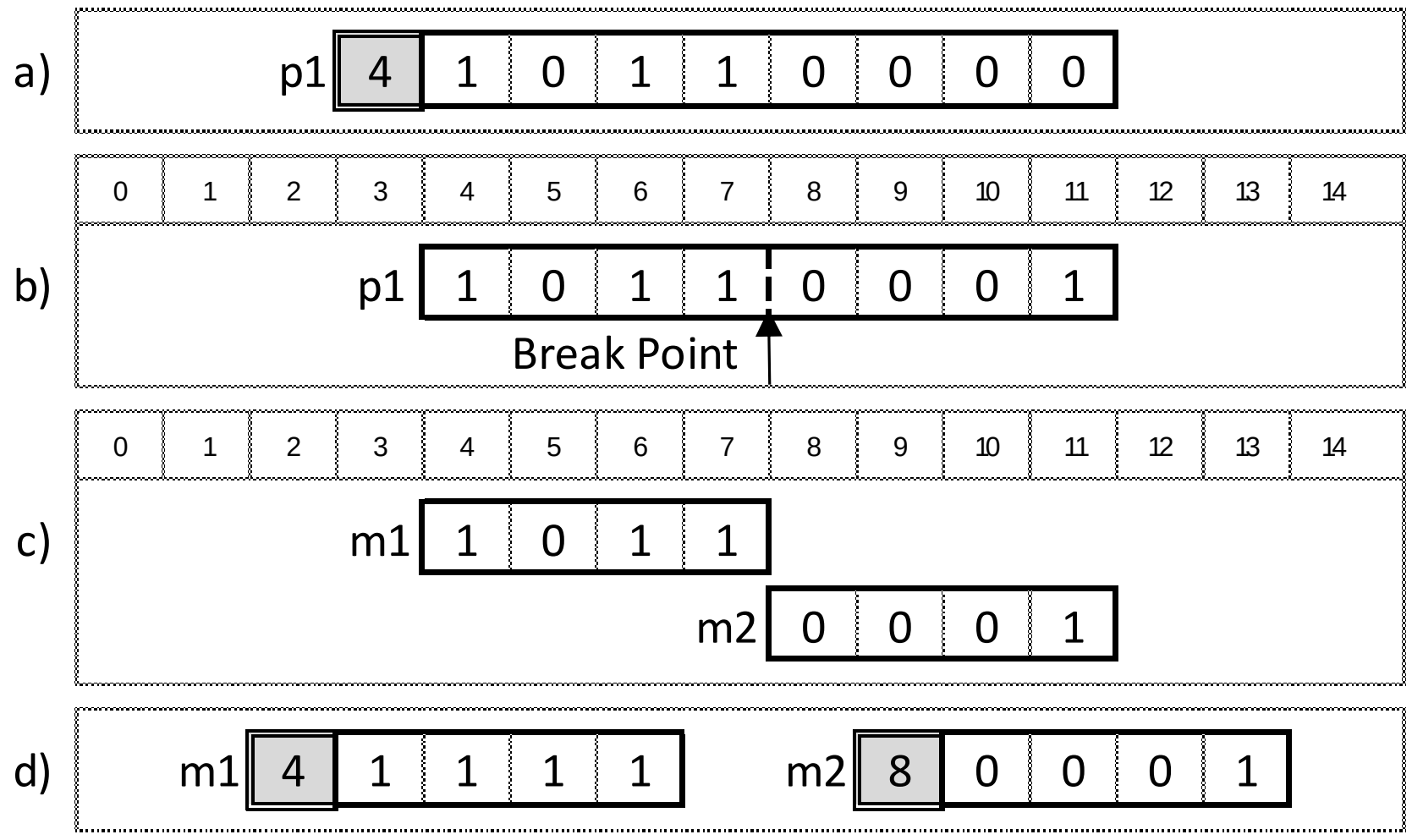

**Fig. 8.** Mutation by breaking genome: a) Original parasite; b) Positions occupied by original parasite and the break point; c) Positions occupied by mutant parasites; d) Mutant parasites.

Equation 1) shows the formula to calculate the probability of parasite splitting. Parameter ***k*** controls the dimension from which the splitting of a parasite is inevitable, i.e when the ratio is greater than one; Parameter ***n*** controls the shape of distribution probability of splitting in other cases. The genotype split point is selected by a uniform probability distribution over the genotype of the parasite.

$$p_{break}(parasite) = \max(\left(\frac{parasite.size^{k}}{host.size}\right)^{n}, 1) \quad (1)$$

This type of mutation avoids disproportionate growth of parasite length and possible subsumption of the host genome. In **Fig. 8** parasite ***p1*** creates two parasites, ***m1*** and ***m2***, where ***m2*** position corresponds to the location splitting point of the parasite genome p1.

## *Evaluation of parasites*

The evaluation of the population of parasites is obtained indirectly through the genomes present in the population of hosts. This feature allows the parasites to be evaluated without the need to apply them to the hosts and then call the fitness function to evaluate the collaboration. In this way, we replace fitness function calls by a proxy consisting of simply checking if the parasite genome is present in the host genome and using the host fitness rank. Therefore, we significantly save function fitness calls as well as computational resources that would be spent on testing and generating collaborations.
We defined three goals for the parasites:

1. Promoting the emergence of parasites with new genetic material, necessary for the evolution of the combined population and prevention of its stagnation;
2. Promoting the dissemination of parasites with good genotypes in the host population so that all individuals have the parasite;
3. Promoting the variability of the anchoring point of good parasites in the host genome in order to allow different regions to be infected.

The last two goals are incompatible with the first since it involves the destruction of the original genetic material. Also, the evaluation function should promote growth of the parasite length to speed up the evolutionary process to discover large BB, and therefore we made the value of parasite fitness directly proportional to its size.

In addition, the evaluation function of the parasites must be independent from the scale of the fitness values in the hosts. To accomplish this, hosts are sorted with a descending rank and parasites use those ranks to compute their evaluation. The parasite evaluation algorithm sums the ranks of the hosts that have the parasite in their genome. If the host rank is defined in the interval [1, n], where n is the rank of fittest host and 1 the worst, parasites that infected the entire population have maximum fitness value. Their contribution to diversification of the population is zero, contrary to goal 1, nevertheless they are good candidates for dissemination, goals 2 and 3. To circumvent this obstacle we shifted the rank of the hosts to the interval [-n/2-1, n/2] where *n* is the size of the population. This shift in ranking of the population provides a number of significant advantages. First of all, the fitness of parasites that infect the entire population is zero;

parasites present only in the best individuals have positive fitness, and by opposition, parasites that are present only in worst individuals have negative fitness.

In order to reward individuals with a large genome, the value of the sum of ranks is multiplied by the size of the parasite. Thus, if a parasite has a positive sum of ranks its size is rewarded; otherwise its size contributes to the decrease of its fitness. Such evaluation makes the discovery of a good parasite to be valuable at the beginning, thereby promoting its spreading, and as it infects the population through successive generations, its interest fades because the population has already assimilated its genome. This parasite evaluation is very efficient because it does not use a single call to the fitness function.

When evolution discovers a new parasite, whose genotype does not exist in the population, the evaluation function should reward its discovery with a fitness that allows it to survive and reproduce if it is a good parasite. On the other hand the length of a new parasite should decrease its fitness to prevent the emergence of large parasites with random genomes that contrast with large parasites evolved from good BB. We decided to assign the new parasite a fitness value equal to the population size divided by its length in bits, as a reward for the discovery of new parasite genomes. The evaluation function allows small parasites with new genotypes to appear in the population and to recombine themselves with existing ones thereby promoting their growth if they contain useful genetic material for evolution.

## *Algorithm of parasites evolution*

The evolution of parasites is done by Algorithm 2. The algorithm receives as parameters the population of parasites to evolve, ***pPop***, the population of hosts to perform the evaluation of the parasites, ***hPop***, and the number of parasites that will be selected to evolve, ***n***.

```
ParasiteEvolution (pPop, hPop, n)
  selectPop = select n parasites from pPop
  offspringPop = recombine selectedPop
  while offspringPop.size < pPop.size
       Select random parasite from offspringPop
       Mutate a clone of parasite
       Insert mutated parasite clone in offspringPop
  End while
  Evaluate offspring in hPop
  pPop = select pPop.size parasites from pPop and
                                        offspringPop
End Function.
```

**Algorithm 2**- Parasite Evolution Algorithm

The algorithm starts by selecting ***n*** parasites from ***pPop***. It continues with the recombination of the selected population giving rise to the population ***offspringPop***. This step recombines genetic material of selected parasites and changes the length of the offspring with the rules described above. The population ***offspringPop*** is constructed by removing a pair of individuals from the selected population, applying the recombination algorithm to the parents and inserting the offspring in ***offspringPop*** population. The algorithm continues completing ***offsringPop*** through successive mutations of clones of randomly selected individuals in offsringPop. One of the three types of mutation described above, genomic mutation, position mutation and genome splitting is randomly applied with uniform probability. This way of completing a population allows a parasite to undergo several mutations in a single generation, because a mutant parasite can be selected and cloned several times.

The population ***offspringPop*** is evaluated through the genes of individuals of the population ***hPop***. The algorithm terminates with the calculation of a new population through replacement operator applied to the original ***pPop*** and to the population of its descendants, the ***offspringPop***.

## *Evolution of hosts*

A population of hosts is evolved with a MuGA algorithm, Algorithm 1, that uses some genetic operators adapted to multipopulations (MP). The adaptation of genetic operators to use the number of copies is critical to MuGA being able to solve difficult problems. MuGA uses standard operators of selection and recombination and an adapted form of mutation and replacement operators. In the next section we describe the adaptions made in operators to take advantages of the number of copies present in MI of MuGA populations.

### MWM - Multiset Wave Mutation

To solve problems where the solution cannot be found by a recombination of parent genes, the mutation operator performs a critical mission to introduce new genes into the population. Mutation introduces random changes in the genome of the individuals. Usually the operator introduces small changes in the genome of the individual and the new features acquired are propagated in the population through generations. A high rate of mutation is required if the changes to escape from local maxima include many alleles but it is harmful if this assumption does not happen. MI in multiset populations represents a set of clones of the same genotype on which we apply different mutation rates.

$$waveFunction(copy) = \left[\frac{\sin(\frac{\pi}{2}+\frac{copy-1}{roughness})}{2}\right]^{thinness} \quad (2)$$

Equation 2 presents a waveFunction formula to calculate the probability of mutation from each clone of the MI that produces values between 0 and 1 (**Fig. 9**). When the mutation value reaches the value 1, all the bits are changed and that feature is very important to optimize deceptive functions where, usually, the optimum is the complement of the local maxima.

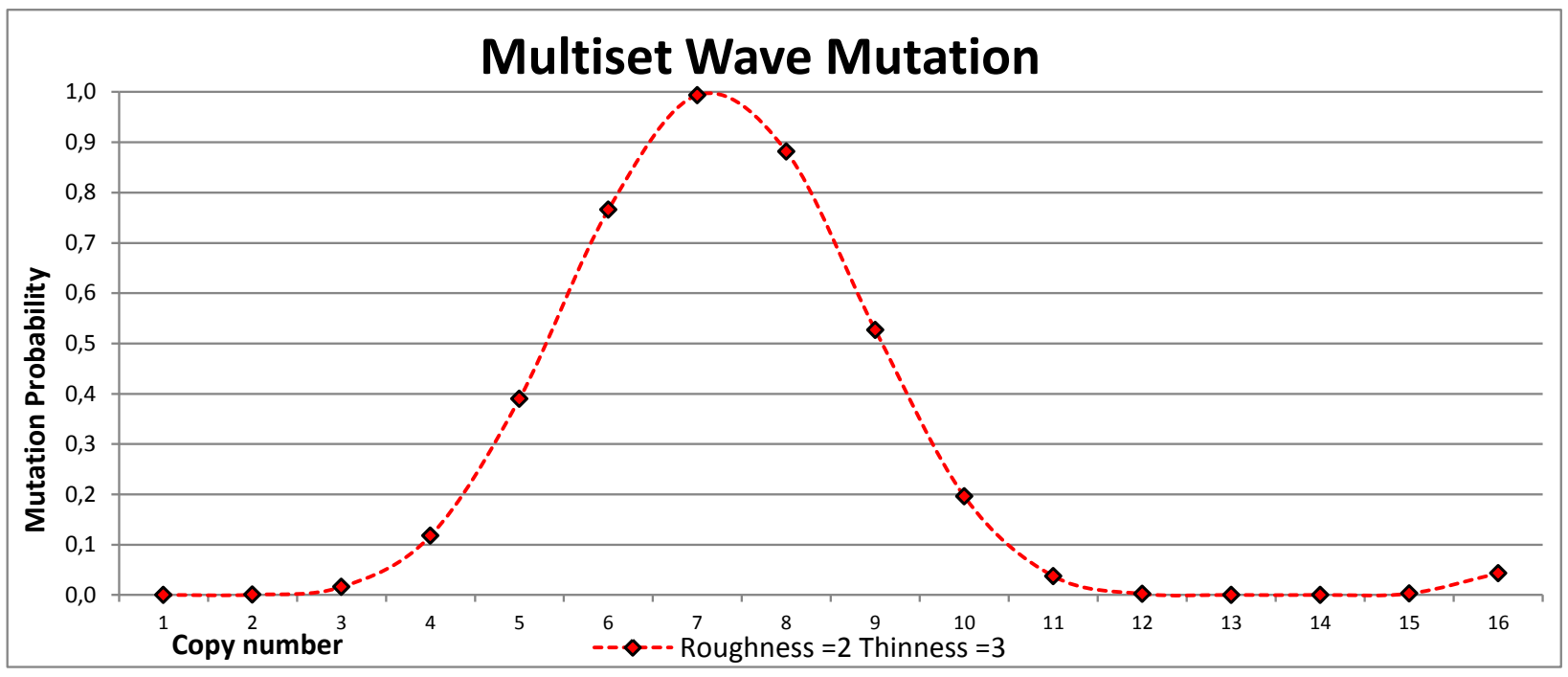

**Fig. 9.** Graph of wave function with roughness=2 and thinness=3

```
MultisetWaveMutation (MI, minProb , mutOperator)
  mutants = empty Multipopulation
  For copy = 1 to MI.copies
    probabilty = min ( minProb + waveFunction(copy), 1)
    individual = MI.genotype
    mutOperator( individual , probability )
    mutants.add( individual )
  next copy
  return mutants
End Function.
```

**Algorithm 3**- Multiset Mutation Algorithm

Multiset Wave Mutation Algorithm 3, fully explained in (Manso and Correia, 2013) was designed to apply a traditional mutation operator, ***mutOperator***, to a multi-individual, ***MI***, using the waveFunction to calculate the probability of mutation of each clone. The probability is calculated adding a minimal probability, ***minProb***, to the result of waveFunction and truncating the result to 1 if the sum is greater than 1. Mutation in the offspring population is brought about by applying Algorithm 3 to every MI present in the population.

## MDR - Multiset Decimation Replacement

The replacement operator has the task of forming the population that will continue the evolutionary process. This operator selects from parents and offspring MP which individuals are selected to continue the evolutionary process.

```
MultisetDecimation (parentsPop,offspingPop , n)
  parentsSize = parentsPop.size
  parentsPop = parentsPop + offspringPop
  while parentsPop.size > parentsSize
      tournament = select n random MultiIndividuals
                                       from parentsPop
      selected = weakest MultiIndividual in tournament
      remove selecteded from parentsPop
  end while
End Function.
```

**Algorithm 4**- Multiset Decimation Algorithm

Multiset Decimation Replacement operator (MDR), Algorithm 4, was designed to replace the parents population with an offspring population in a steady state approach maintaining the multiset characteristics of MI present in both populations. MDR joins the offspring population with the parents population and the individuals with the same genotype increase their number of copies. The algorithm then selects a group of random MI and removes the weakest. This procedure is repeated until the parent population is reduced to the same number of MI of the original population.

## *Co-evolution of hosts and parasites*

The algorithm SMuGA is an evolutionary algorithm that uses two cooperating populations to solve difficult problems: the host population that contains solutions of the problem, and the parasite population that helps the first to reach the best solution. Parasite populations evolve to achieve good genes that represent partial solutions, and infect hosts through the incorporation of those genes.
The interaction between hosts and parasites produces a new population using symbiosis that mimics what occurs in the natural world. We define collaboration as the result of a host infected by one or more parasites using symbiosis.

### Collaboration

A collaboration is obtained by copying the alleles of the parasite into the host. In this case the alleles of the host are replaced by those of the parasite.

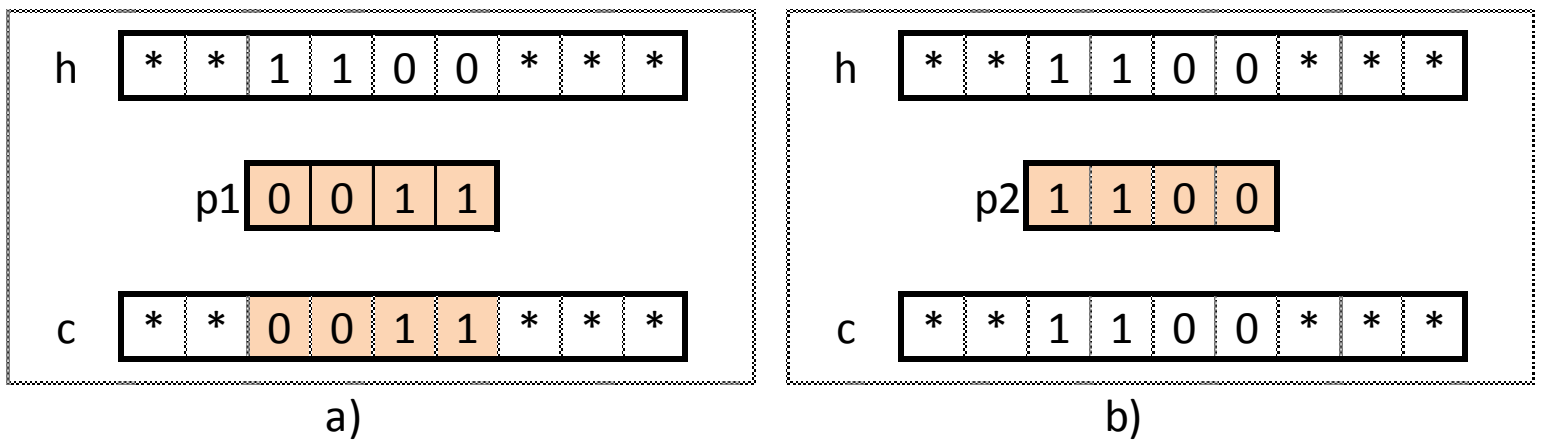

**Fig. 10.** Collaboration between one host and one parasite: a) successful collaboration b) collaboration rejected.

A collaboration of a parasite with a host is only allowed if the host does not have all the bits of the parasite, **Fig. 10** a). This means that a parasite can infect a host only once, **Fig. 10** b). This detail allows the elimination of collaborations that do not add anything new, and clears space for collaborations that do modify something in the host.

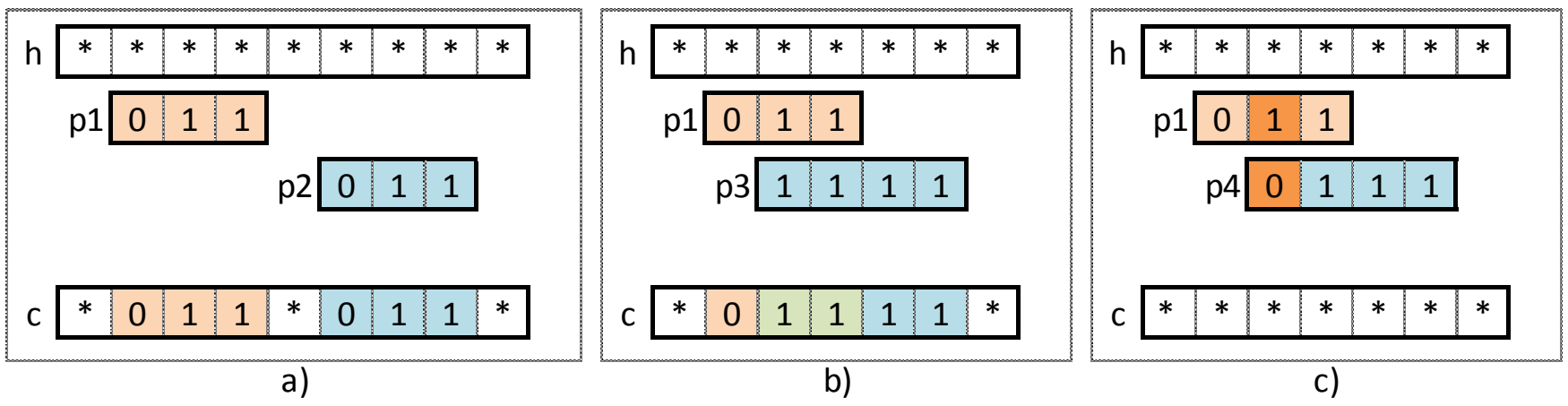

**Fig. 11.** Infection of a host by two parasites: a) non overlapping parasites, b) compatible overlapping parasites c) incompatible overlapping parasites

We restrict the application of multiple parasites to cases where parasites do not have incompatible bits. This means that the parasites may overlap, provided that the overlapping segment does not contain different bits.

In **Fig. 11** a) parasites ***p1*** and ***p2*** can infect host ***h*** because they infect disjoint regions. **Fig. 11** b) parasites ***p1*** and ***p3*** can infect the host ***h*** because, although they share two genes, they have the same value and therefore the infection causes no ambiguity. In **Fig. 11** c) parasites ***p1*** and ***p2*** cannot be used simultaneously because they overlap in two genes, one of which has distinct alleles. In this case, the host can be infected by any of them but not by both simultaneously.

Algorithm 5 controls the formation of collaborations among a population of hosts and a population of parasites. Algorithm 5 takes as parameters a host multipopulation, ***sortedHostPop***, sorted in descending order, a parasite population, ***parasitePop*** and a parameter ***n*** that controls the probability of infection. The order of the population is important because the index of the host in a population determines the probability of the host receiving parasites. The algorithm continues with the definition of the population resulting from the collaboration, ***symbPop***, among populations that are passed as a parameter. Afterwards, the hosts are selected sequentially and the probability of infection is calculated. As hosts are MI, the algorithm proceeds to expand into clones and applies parasites to each one of them independently. Individuals with a higher ranking are those that usually make more copies and thus may suffer various combinations of parasites.

After selecting a host and calculating a probability of infection, the algorithm continues with the application of parasites to each of its clones. The parasites are randomly arranged within the population of parasites to ensure no preference in its application. In the next step the algorithm tries to apply each parasite to the host selected using the compatibility rules of **Fig. 11**. In order to preserve the good individuals of the population from a generalized infection, and hence the sudden change of its genome, parasites are applied in a probabilistic manner. A host is particularly vulnerable to parasites when its rank in the population is smaller. This allows the fittest individuals to receive few parasites, thereby preserving their genes, and lower-ranked individuals are subject to a generalized infection accommodating several parasites. This process is similar to that described in (Dumeur, 1996).

$$p_{infection}(h, pop) = \left(\frac{rank(h,pop)}{pop.size}\right)^n \tag{3}$$

Equation 3) shows the formula to calculate the probability of a parasite infecting a host, ***h,*** contained within a population, ***pop***. The ***rank*** function returns the rank of the individual within the population, in descending order of fitness and ***pop***.size represents the number of hosts that the population has. Parameter ***n*** controls the shape of the ratio described above.

```
Collaboration (parasitesPop, sortedHostsPop, n)
  symPop = empty MultiPopulation
  for index = 1 to sortedHostsPop.size
    host = sortedHostsPop.get(index)
    pInfection = (index / hosts.size)^n
    for copy = 1 to host.numberOfCopies
       symbiosis = host.genotype
       randomize parasites in parasitesPop
       foreach parasite in parasitesPop
          if compatible(parasite, symbiosis) and
                  uniformRandom(0,1) < pInfection
             symbiosis = symbiosis + parasite
             add symbiosis.clone to symbPop
          end if
       next parasite
    next copy
  next index
  return symbPop
End Function.
```

**Algorithm 5**- Collaboration between Hosts and Parasites

The symbiosis population is built by the infection of selected parasites into the host genomes. When a parasite is applied to the host, the genome of the parasite is copied to the genome of the host generating a new individual through symbiosis. A clone of that collaboration is added to the population of symbiosis, and the symbiosis continues the process of being infected by other parasites.

## SMuGA - Symbiogenetic Multiset Genetic Algorithm

SMuga, Algorithm 6, uses multipopulations to represent the populations of hosts and parasites. This representation enables the use of multiset-adapted genetic operators in both populations to help the evolutionary process. The use of multipopulations is required to optimize deceptive problems, and every challenging problem has a degree of deception (Whitley, 1991). This algorithm has two phases: the collaboration phase, where the parasites infect the hosts; and the evolution phase, where hosts and parasites evolve using coevolution.

```
SMuGA (h, p, problem, iterations, k, n)
  hPop = generate h MultiIndividuals from problem
  Evaluate hPop
  pPop = generate p MultiParasites from problem
  Evaluate pPop with hPop
  Repeat
     /* Collaboration phase */
     selPop = Select k hosts from hPop
     symbPop = collaboration( pPop, selPop, n)
     hPop = Select h hosts from symbPop and hPop
     /* Evolution phase */
     Repeat iterations times
        Evolve hPop
        Evolve pPop
     End repeat
  Until stop criteria
End Function.
```

**Algorithm 6**- SMuGA – Symbiogenetic Multiset Genetic Algorithm

The algorithm has six parameters: ***h*** represents the size of the host population; ***p*** the size of the parasite population; ***problem*** the problem to be solved; ***iterations*** the number of iterations that hosts and parasites evolve without collaboration; ***k*** the number of hosts selected to participate in the collaboration; and ***n*** that controls the probability of hosts infection.

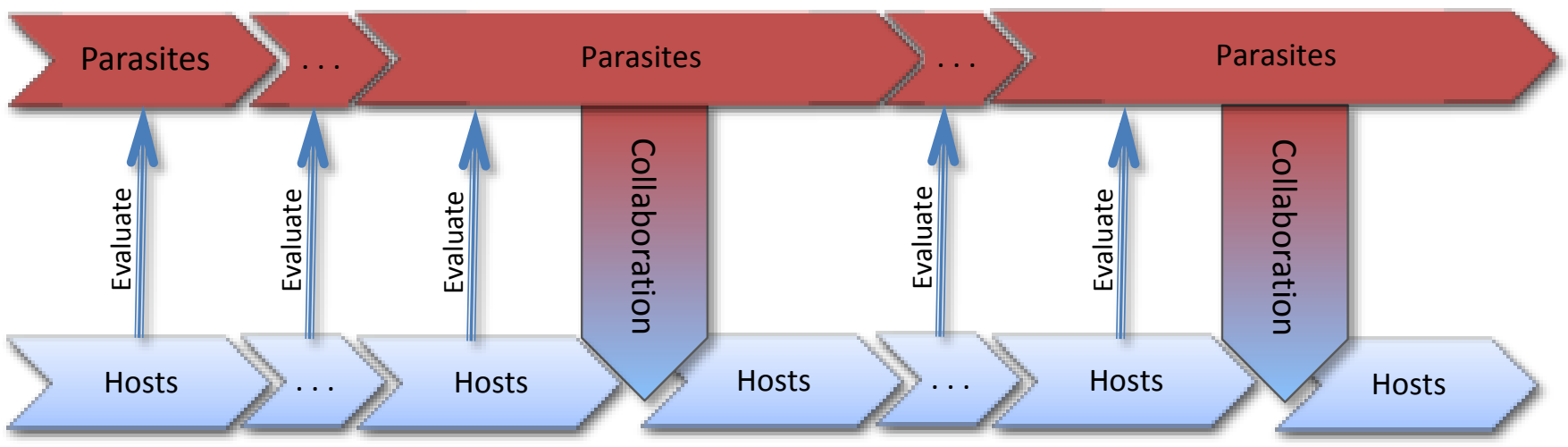

**Fig. 12.** Interaction between populations in SMuGA algorithm.

The algorithm starts by generating and evaluating a host population, ***hPop***, with ***h*** hosts of ***problem***, and a parasite population, ***pPop***, with ***p*** parasites. The only information about the problem needed by parasites is the size of the host to perform mutations. The evaluation of ***pPop*** is done using ***hPop***. **Fig. 12** show the interaction between ***sPop*** and ***hPop***.

The evolutionary iterative process starts with the collaboration phase followed by the evolution phase until a stop criteria is reached.

Collaboration phase is performed by Algorithm 5 between populations of parasites, ***pPop***, and the ***k*** selected hosts in the host population, ***selPop***, using the parameter ***n*** to control the infection probability of hosts. The result of Algorithm 5 is a symbiosis population, ***symbPop*** that contains the selected host clones infected by the parasites. Because one host may be infected by many parasites and the algorithm saves clones when a host is infected by one parasite, the number of symbiosis is huge when compared to the number of parasites and number of hosts. This phase is computationally expensive. That effort is relieved by the use of multipopulations since the collaboration algorithm produces symbiosis with repeated genotypes and the multiset representation helps in its storage and evaluation. The collaboration phase ends with the selection of ***h*** hosts from the union of host population, ***hPop*** and symbiosis population, ***symbPop***.

The evolution phase starts with the evolution of ***hPop*** using MuGA algorithm, Algorithm 1, and the evolution of ***pPop*** using Algorithm 2. Both populations evolve for ***iteration*** generations without establishing new collaborations. This phase is used to stabilize the individuals in the populations and to assimilate, in the hosts, genetic material introduced by the collaboration phase. The host population evolves on its own, however, the parasite population still uses hosts, since parasites are evaluated using the genes of the host population as a proxy for fitness evaluation. When hosts evolve and change their genes, the fitness value of parasites may change too.

## Experimental study

To examine the influence of symbiosis in the solutions of hard problems we conducted a set of experiments with the SMuGA algorithm and compared the results with the standard MuGA. We compared, also, the results of SMuGA with SCA in order to assess the scalability of the algorithm to big deceptive problems.

## *Experimental Setup*

MuGA was configured with 128 MI in the main population. Selection is made by tournaments with size 3. The operator selects 256 individuals for the mating pool and in this way MI with copies are guaranteed for the following operators. Recombination is made by one point crossover operator with probability 0.6. Mutation is made by the multiset wave mutation, MWM, configured with roughness 2 and thinness 3 (**Fig. 9**). The minimal mutation probability, parameter ***minProb*** of Algorithm 3, is equal to ***1/l*** , where ***l*** represents the size in bits of the genome of the individual. Rescaling was applied to maintain a maximum total of copies in the main population of twice the number of MI.

SMuGA is configured with 32 MI in the host population and 32 MI in the parasite Populations. In this case, we can use a smaller population than with MuGA, due to the increased genetic variety introduced by parasites. The size of the population selected to make collaboration is 16 MI, and the parameter that controls the probability of infection, parameter ***n*** in Algorithm 5, has value 1. The number of iterations of the evolution phase in Algorithm 6 is set to 16. The evolution of hosts uses tournament selection with tournament size 3 and selects 32 individuals. Recombination is done by uniform crossover with probability 0.6 and mutation, replacement and rescaling are performed in the same way as in MuGA. **Table 1** shows evolutionary parameters of MuGA and SMuGA.

**Table 1**- Configuration of MuGA and SMuGA

| | **MuGA** | | **SMuGA** | |
|---|---|---|---|---|
| | Parameter | Settings | Parameter | Settings |
| Size of Population | Individuals | 128 | Hosts | 32 |
| | | | Parasites | 32 |
| Selection | Tournament size 3 | 256 | Tournament size 3 | 32 |
| Recombination | Crossover 1 cut | 0.6 | Uniform Crossover | 0.6 |
| Mutation | MWM | 2 , 3 | MWM | 2 , 3 |
| Replacement | Decimation | 2 | Decimation | 2 |
| Rescalling | Adaptive | 2 | Adaptive | 2 |

To obtain statistical confidence we performed 128 independent runs for each experiment. In each run, random initial populations were generated for individuals, hosts and parasites. The stop criteria used in the simulations is the number of evaluation function calls, and due to the varied difficulty of the problems that limit is adjusted to allow the success of the evolutionary process. For each experiment we compute the average of the number of evaluations to find the optimum. We assign the maximum number of evaluations to the experiments where the optimum is not found. We also compute what we consider a more revealing result, which is the success rate, meaning the percentage of runs that reach the optimum.

To compare the algorithms we use pair-wise Student T tests with 95% confidence interval for the means. Due to the large number of simulations we assume the normality of the variables. For each problem we also compare results with other previously referred algorithms, when available, which means only for smaller genome lengths. However, results published for these problems are not always precise. In some cases only logarithmic graphs are printed and the results here presented are best effort readings. And they never present the percentage of runs that reach the optimum.

## *Experimental results with deceptive functions*

The key to the success of Evolutionary Algorithms (EA) is their combination of low-order building blocks (BB) to form higher-order BB, which eventually leads to the optimum. When the solution cannot be built through this incremental combination of BB, we are in the presence of deceptive problems and we need to improve the artificial evolutionary process in order to solve those problems. The concept of deception was first introduced by Goldberg (Goldberg, 1987) and much work has been done in addressing this class of problems. MuGA and SCA are two evolutionary algorithms that are able to optimize deceptive functions. In the next sections we present experimental results on different deceptive benchmark functions, for SMuGA, MuGA and SCA.

### Fully deceptive F3 Function

Goldberg (Goldberg, 1989), devised a 3-bit function, F3, presented in Equation 4, that is fully deceptive since building blocks of order n are deceptive to build blocks of order n+1.

$$F3(000) = 28,\quad F3(001) = 26,\quad F3(010) = 22,\quad F3(011) = 0 \tag{4}$$

$$F3(100) = 14,\quad F3(101) = 0,\quad F3(110) = 0,\quad F3(111) = 30$$

Fully deceptive function ***F3*** is easily solved by EA due to is size of three bits. To get a changeling problem we define the function ***F3 10*** as ten consecutive copies of ***F3***. This procedure is usual in the optimization in this kind of deceptive problems and is adequate to be solved using symbiogenesis present in SMuGA.

Optimization of ***F3 10*** was successfully solved by the two algorithms (SMuGA, MuGA), **Table 2**, and the symbiotic approach speeds up the evolutionary process. **Fig. 15** shows the evolution of the success rate of the algorithms in the first 30,000 evaluation function calls, and **Fig. 14** presents a statistical view of the number of evaluation function calls needed to reach the optimum in both algorithms. Results of SMuGA in function ***F3 10*** are more than on order of magnitude better than those presented by (Yang, 2004) and (Chen et al., 2008).

**Table 2**- Statistics of SMuGA and MuGA result in F3 10 function.

| | SMuGA | | MuGA | |
|---|---|---|---|---|
| F3 10 | Mean | Std | Mean | Std |
| Evals. to find Best | 3309.79 | 1273.36 | 6074.30 | 2516.68 |
| Best value found | 300.00 | 0.00 | 300.00 | 0.00 |
| Sucess rate (%) | 100.00 | 0.00 | 100.00 | 0.00 |

**Fig. 13** shows in more detail the evolution of the success rate, observing only the first 6,000 evaluation function calls. In that figure we can clearly see, in the major steps, the effect of the periodic incorporation of parasites in hosts, when new collaborations are formed and integrated into the host population. The evolution of the isolated host population over a few generations allows spreading of good genetic material introduced by symbionts through the population. The parasite population evolves in parallel, in this case taking into account the host population to estimate the fitness of parasites. This process is very economical in the number of collaborations generated, and subsequent calls to the fitness function.

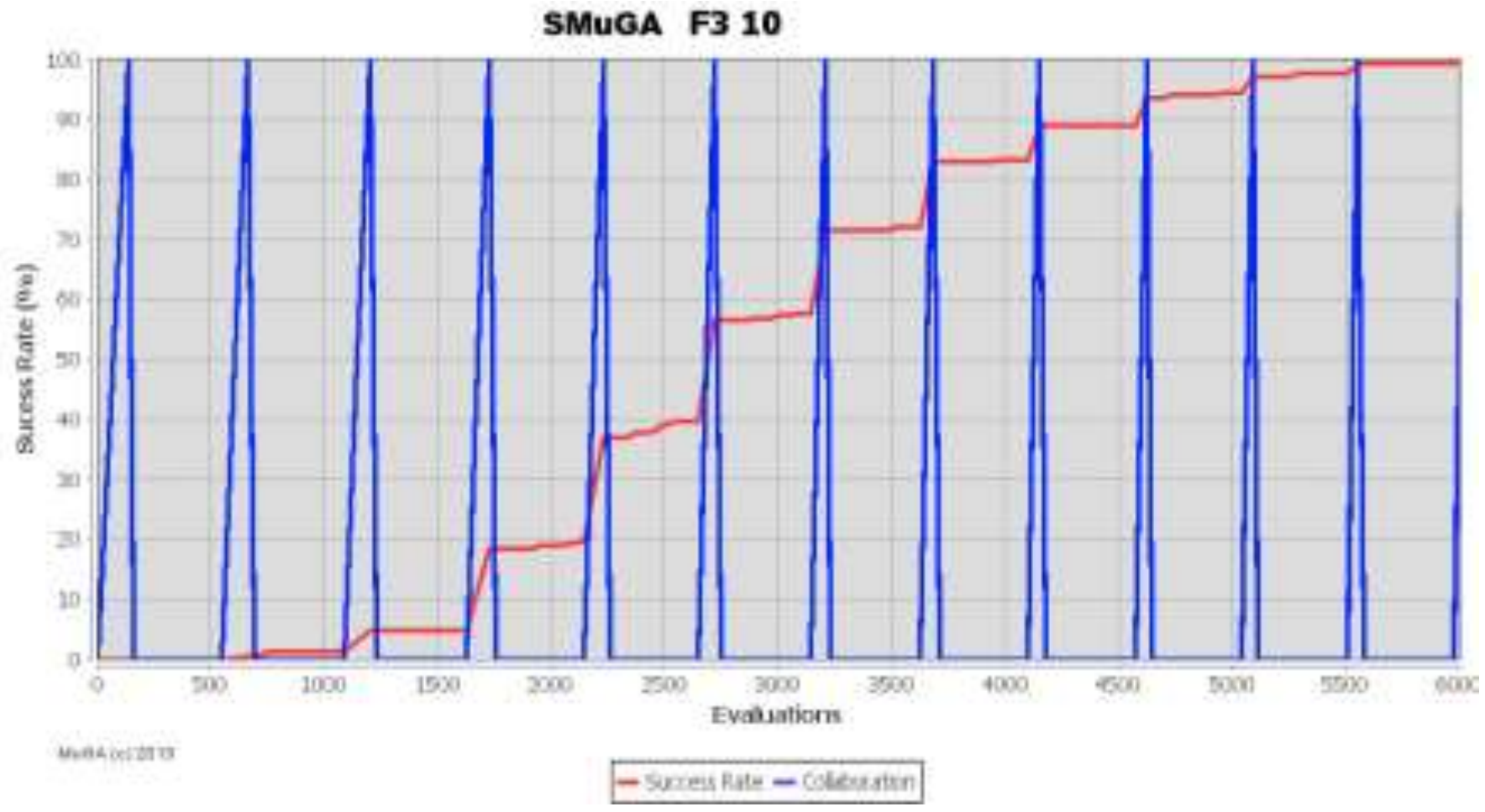

**Fig. 13.** Detail of the evolution of the success rate in the optimization of 10 copies of F3 Function with SMuGA solver. Blue line represents the collaboration event between hosts and parasites.

Function ***F3 10*** is solved by SMuGA due the use of symbiosis between hosts and parasites. If one parasite that represents a BB of the function is found, it may be copied to the position where another BB starts and the fitness of the collaboration is sharply increased. The search for the BB and their positions is not easy because no information about the function landscape is provided to SMuGA. Remarkably SMuGA finds adequate length BB and their positions and uses symbiosis in a very efficient way.

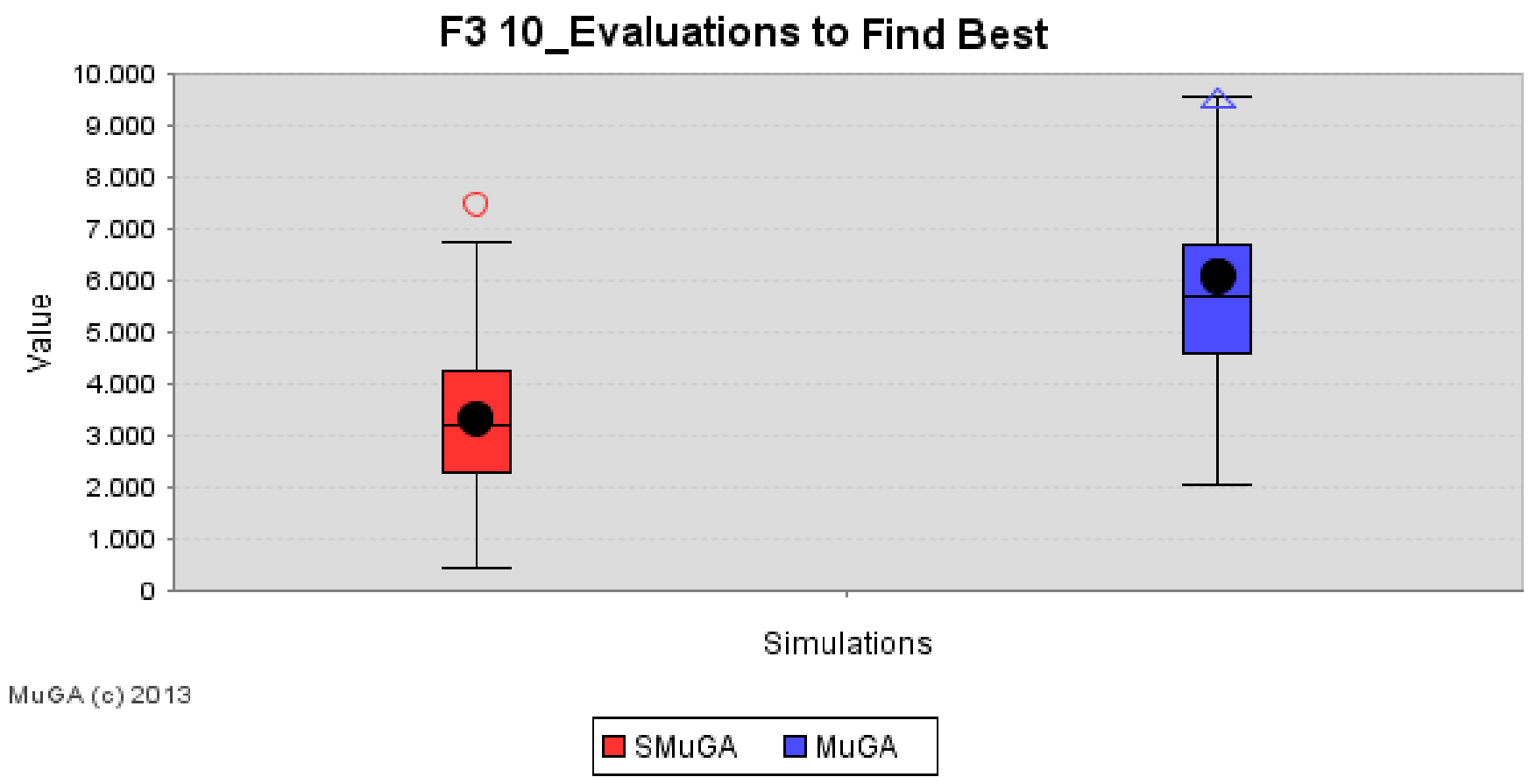

**Fig. 14.** Box-plots of the evaluation function calls to find the best of 10 copies of F3 Function

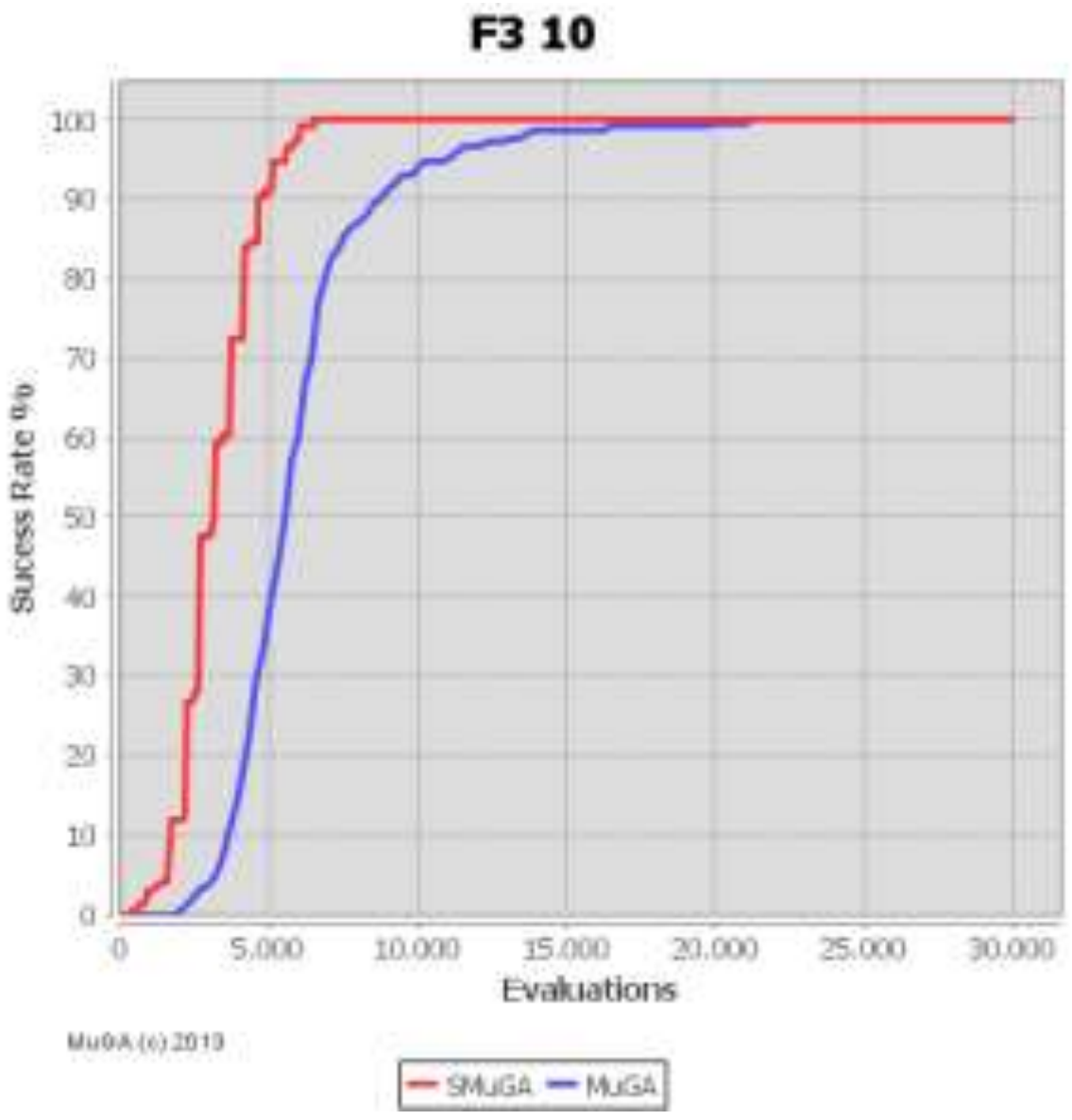

**Fig. 15.** Evolution of the success rate in the optimization of 10 copies of F3 Function.

**Table 3**- Statistics of SMuGA evolution result in optimization of concatenated F3 function with different lengths.

| SMuGA | Evals. to find best | | Sucess % | |
|---|---|---|---|---|
| F3 | Mean | Std | Mean | Std |
| 30 bits | 3088.69 | 1363.68 | 100.00 | 0.00 |
| 60 bits | 5054.30 | 1495.08 | 100.00 | 0.00 |
| 120 bits | 8457.08 | 3021.80 | 100.00 | 0.00 |
| 240 bits | 17500.25 | 12182.22 | 98.44 | 12.40 |
| 480 bits | 24960.44 | 13143.26 | 95.31 | 21.14 |

In order to verify the scalability of the algorithm SMuGA to big genome problems we performed a set of tests with the composition of 10, 20, 40, 80, and 160 fully deceptive F3 functions corresponding to problems with 30, 60, 120, 240 and 480 bits respectively. For these tests we only present results for SMuGA since, in large problems, MuGA does not achieve solutions in reasonable time, and other algorithms do not present results.

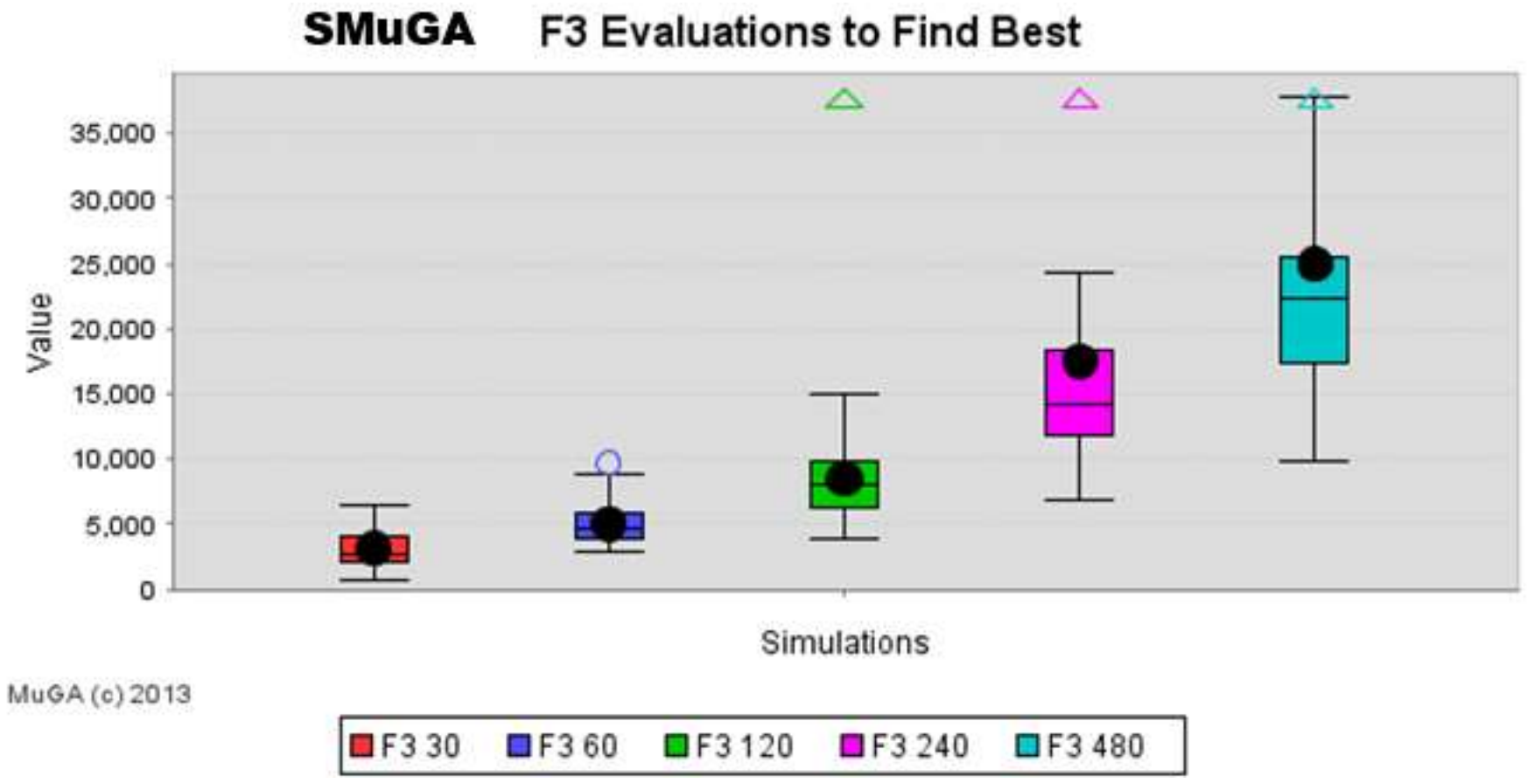

**Fig. 16.** SMuGA: Box-plots of the evaluation function calls to find the best value in 10(30), 20(60), 40(120), 80(240) and 160(480) copies(bits) of F3 function. Notice that vertical axis is linear while horizontal axis is exponential.

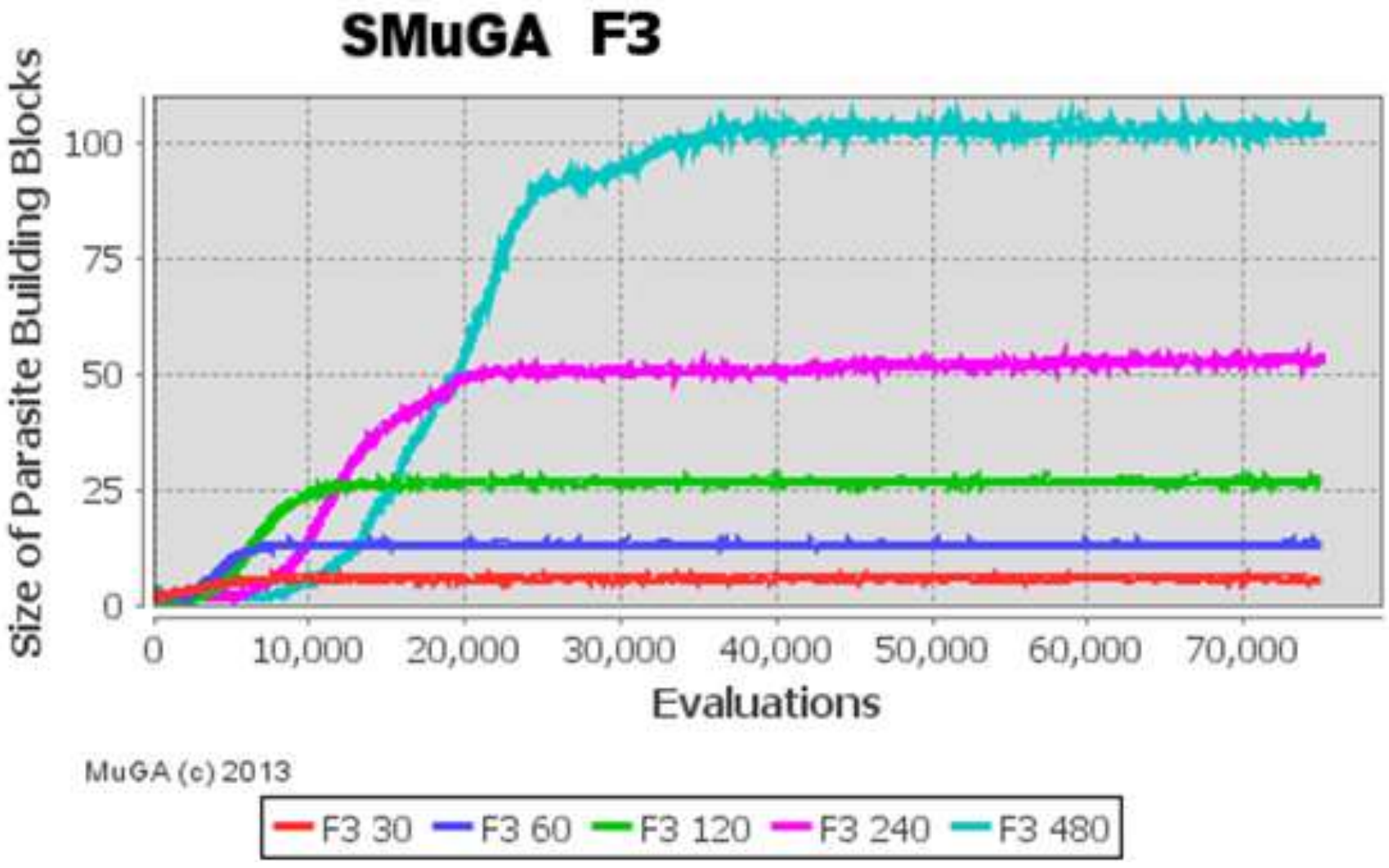

**Fig. 17.** SMuGA: Evolution of the size of building blocks of 10, 20, 40, 80 and 160 copies of F3 function

**Table 3** and **Fig. 16** show the evolution statistics in the optimization of the concatenated F3 function with different lengths using SMuGA after 75,000 function evaluations calls. The algorithm scales in a linear way in this kind of functions due to its ability in finding good BB, assembling them with recombination, **Fig. 3**, and thus forming larger BB which can be moved to other locations in the genome, **Fig. 6**. This feature allows the solution of problems with long genomes of concatenated functions in a very efficient way. **Fig. 17** shows the evolution of the size of building blocks in that experiment. As we can see, problems with long genomes are solved by parasites also with long genomes, which will eventually incorporate a collaboration, speeding up the evolution of hosts. Again we note that the algorithm does not receive any information about BB.

**Fig. 18** shows the evolution of the success rate. The decrease of success in optimization of F3 with 240 bits, 98%, and 480 bits, 95 %, can be explained by the small size of the parasite population (32 individuals) for a very large genome of the hosts. In that case, the probability of assembling useful BB in parasites decreases due to the large space that they explore.

For the functions analyzed next, we notice similar behavior to the one depicted in **Fig. 13** in the step growth of success rate; and also a similar result to the one depicted in **Fig. 17**, regarding the evolution of the length of parasites as a function of the size of the problem. Therefore, we do not present such graphs for the remaining functions.

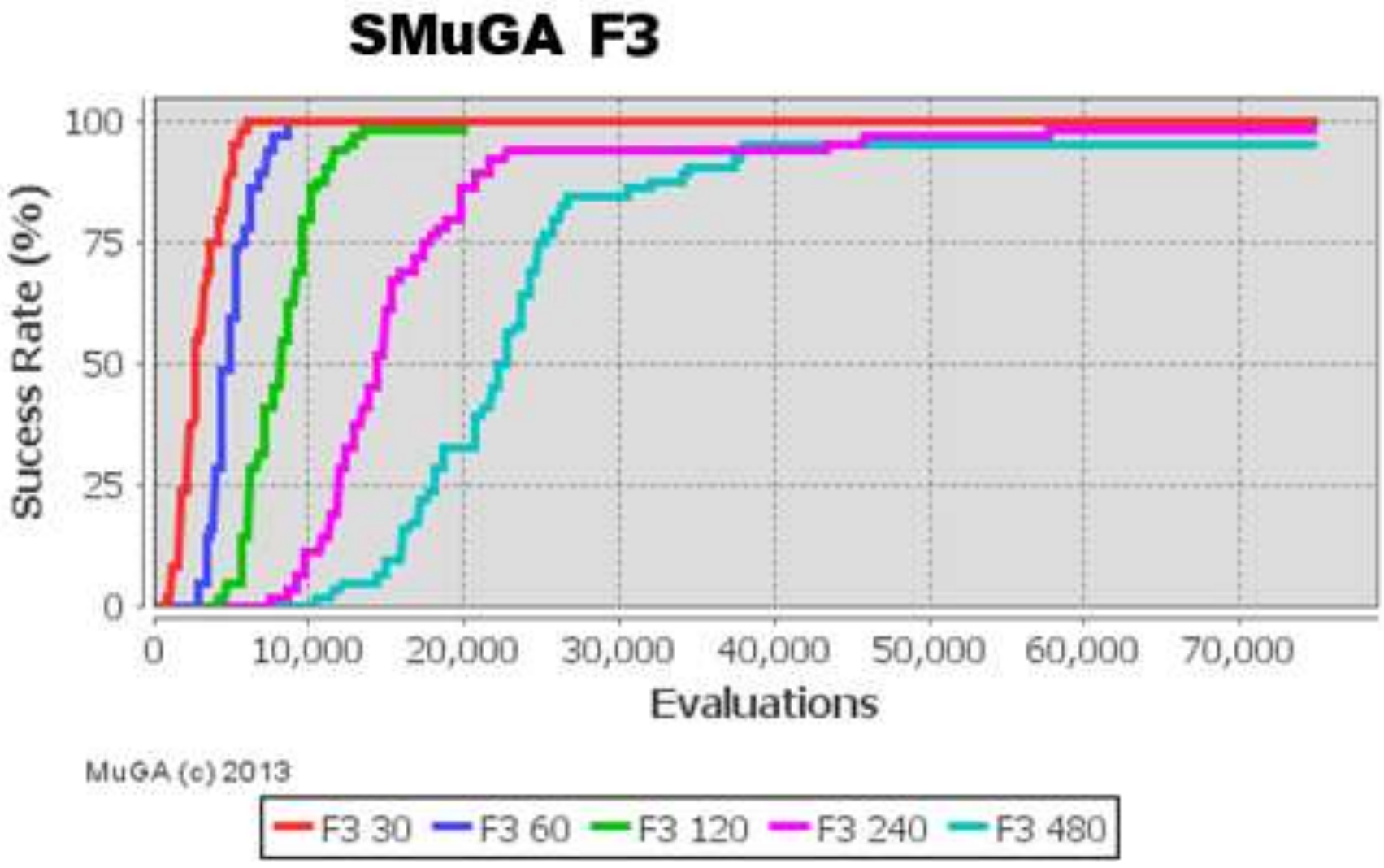

**Fig. 18.** SMuGA: Evolution of the success rate in the optimization of 10, 20, 40, 80 and 160 copies of F3 function

### Maximally separated fully deceptive F3 Function

The composition of functions in a sequential way is solved by SMuGA using the mobility property of parasites present in the algorithm. The application of one good parasite, which represents a BB, in a position where other BB starts, contributes to the success of the algorithm due to the nature of the function composition.

The problem becomes difficult when the bits of each function are separated. The most difficult case of separation is when they are uniformly and maximally distributed in the chromosome. We call these functions ***F3S N***, where ***N*** represents the number of F3 functions in the chromosome. In case of ***F3S 10*** each bit of one function is located in positions ***i***, ***i+10***, and ***i+20***.

**Table 4**- Statistics of SMuGA and MuGA result in F3S 10 function

| | SMuGA | | MuGA | |
|---|---|---|---|---|
| F3 Separated | Mean | Std | Mean | Std |
| Evals. to Find Best | 9419.20 | 5604.92 | 34063.77 | 18018.62 |
| Best value found | 300.00 | 0.00 | 299.95 | 0.30 |
| Sucess rate (%) | 100.00 | 0.00 | 99.22 | 8.80 |

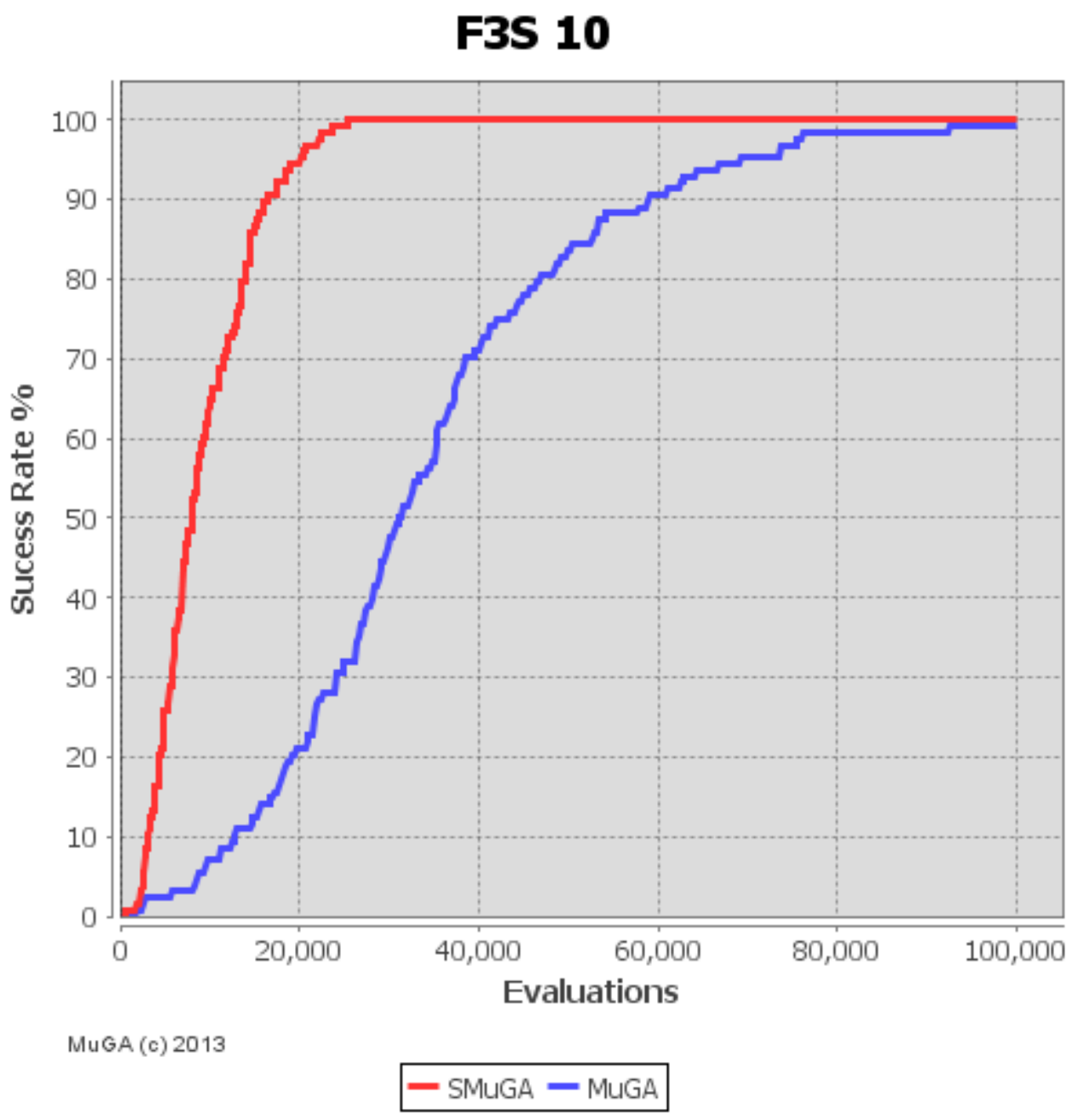

**Fig. 19.** Evolution of the success rate in the optimization of F3S 10 function

These functions are difficult because the problem is not separable and the formation of BB is not possible with a naïve strategy. In this way the bits of the functions are spread and the application of one parasite in different positions is not enough to solve the problem. SMuGA escapes this situation by combining several parasites in a single host. With this experiment we verify SMuGA’s effectiveness in non-separable problems as well.

**Table 4** presents the results of the optimization of ***F3S 10***. Both SMuGA and MuGA solve the function with notable efficacy and, again, symbiogenesis speeds up the evolutionary process. **Fig. 19** shows the evolution of the success rate of the algorithms in the first 100,000 evaluation function calls, and **Fig. 14** presents a statistical view of the number of function evaluations needed to reach the optimum in both algorithms.

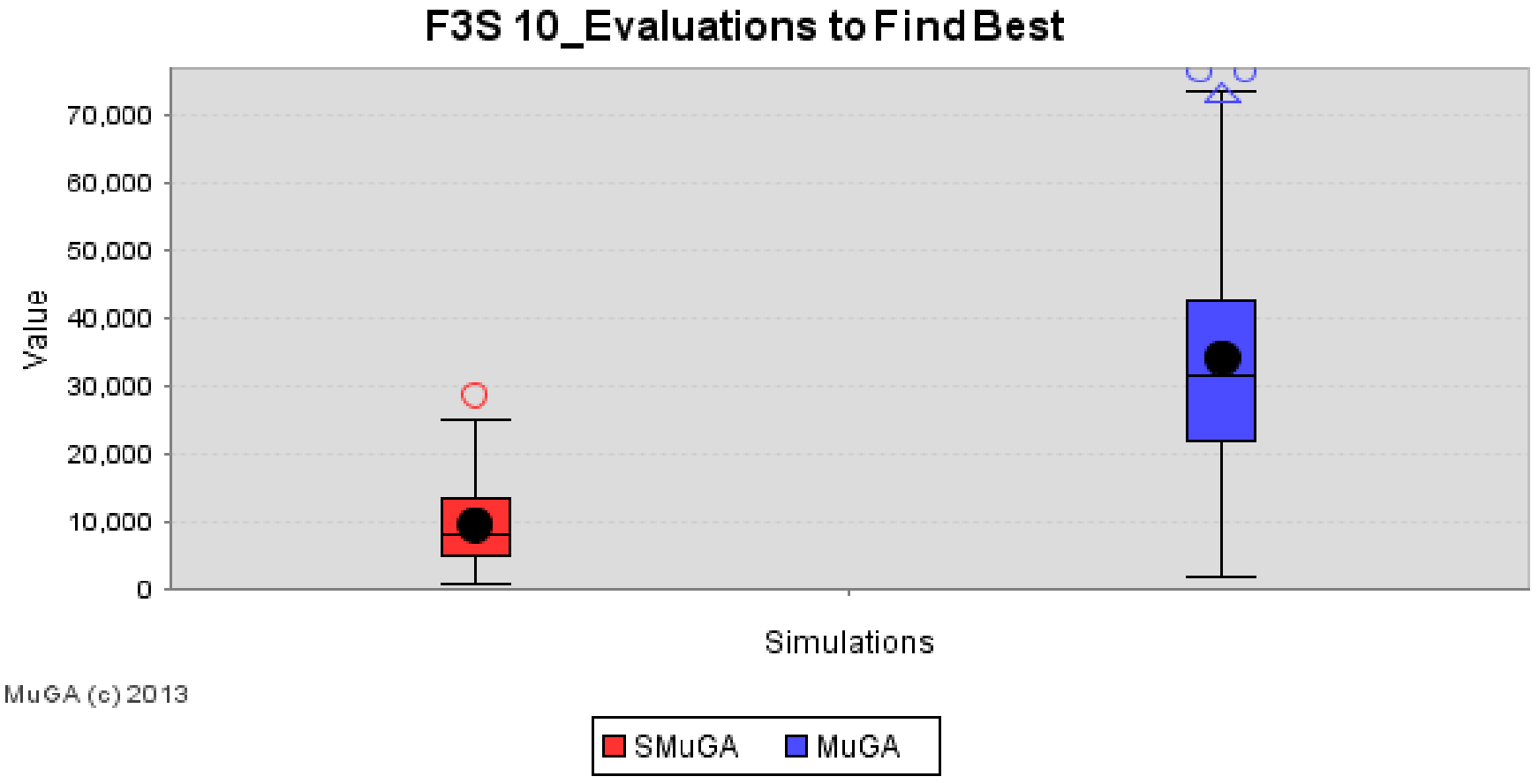

**Fig. 20.** Box-Plot of the evaluations to find the best in the optimization of F3S 10 function

**Table 5**- Statistics of SMuGA evolution result in optimization of separated F3S with different lengths.

| **SMuGA** | Evals. to find best | | Sucess % | |
|---|---|---|---|---|
| **F3S** | Mean | Std | Mean | Std |
| 30 bits | 7651.08 | 5049.29 | 100.00 | 0.00 |
| 60 bits | 26429.98 | 19408.04 | 100.00 | 0.00 |
| 120 bits | 184549.02 | 145359.55 | 95.31 | 21.14 |
| 240 bits | 381741.92 | 148172.39 | 51.56 | 49.98 |
| 480 bits | 496565.75 | 27516.57 | 1.56 | 12.40 |

**Fig. 21** and **Table 5** shows the statistics of the optimization of the composition of 10, 20, 40, 80 and 160 F3S function in the chromosome after 500,000 evaluation function calls. As previously stated, the bits of F3S N functions are maximally spread over the chromosome, and big genomes separate the bits of one function with large distances. SMuGA fully succeeds in the optimization of 10 and 20 F3S N functions. In the optimization of 40 F3S, whose chromosome has 120 bits and the bits of each F3S function are separated by 40 bits, SMuGA succeeds in 95% of simulations and needs more generations to fully succeed. In the larger simulations, the small population of parasites and the large genome of the hosts hinder the optimization, and the parameters must be adjusted.

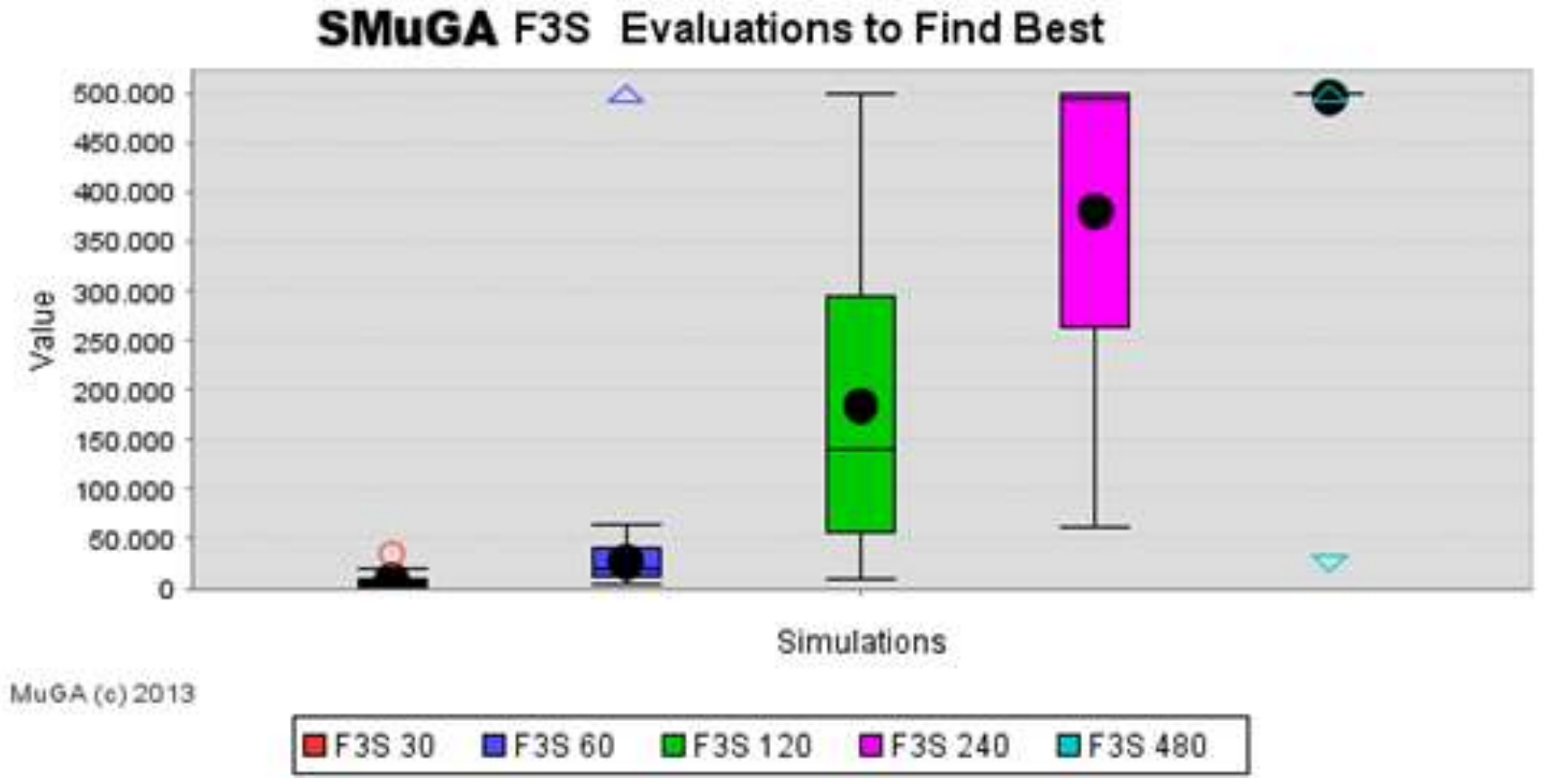

**Fig. 21.** SMuGA: Box-plots of the evaluations to find the best value in 10, 20, 40, 80 and 160 copies of F3S function.

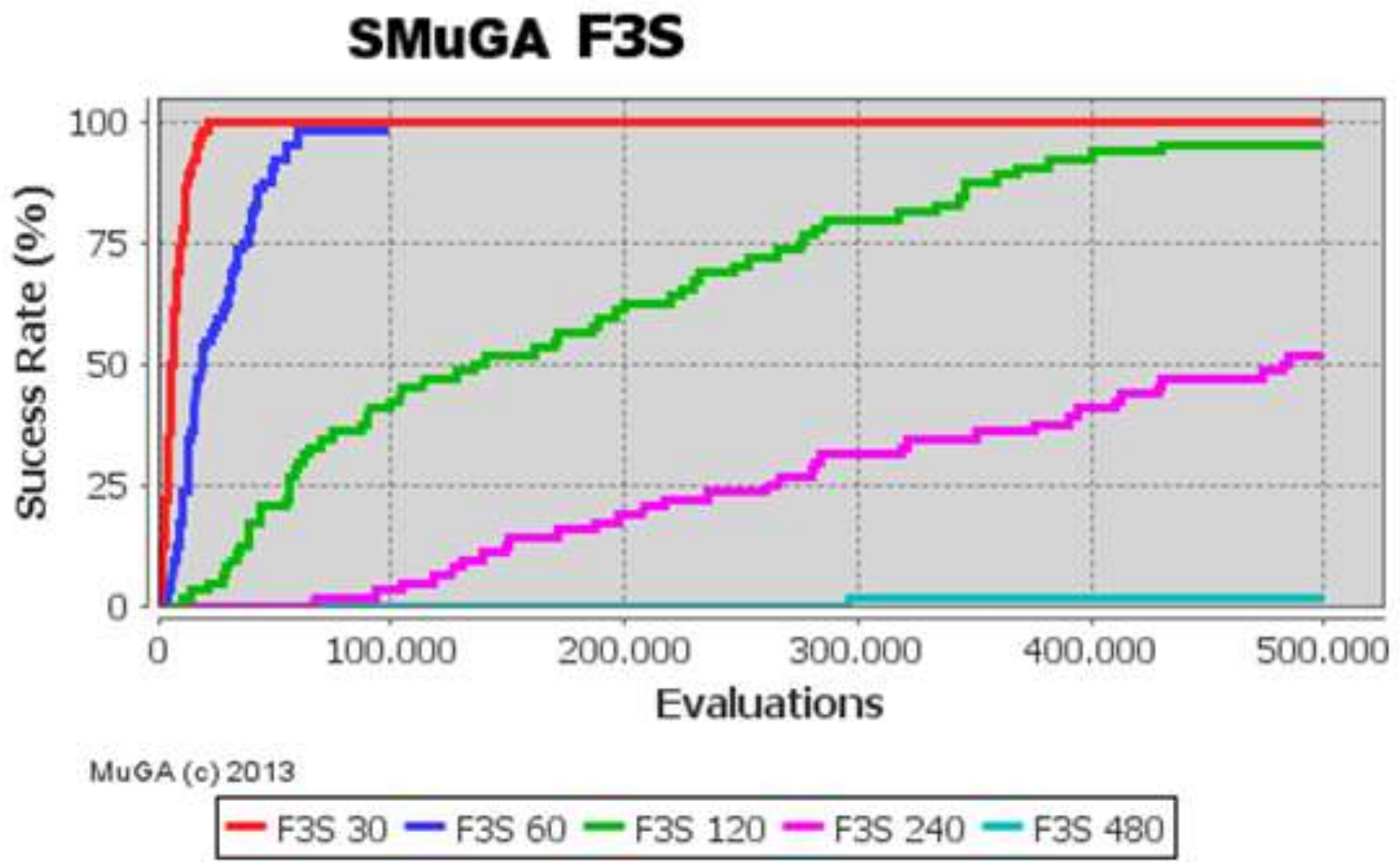

**Fig. 22.** SMuGA: Evolution of the success rate in the optimization of 10, 20, 40, 80 and 160 copies of F3 function.

## Deceptive functions

Deceptive functions, also referred to as trap functions, were introduced by Ackley (Ackley, 1987) and are defined in the unitation space. In this space, only the number of ones in the chromosome counts, regardless of the order. Equation 5 presents the formula of a deceptive function where ***x*** is the chromosome, ***u(x)*** is the number of ones in the chromosome ***x*** and ***l*** represents the length of chromosome ***x***. **Fig. 23** presents a deceptive function with four bits in the unitation space. This allows us to test the algorithm with a larger function and for which there are other models with published results.

$$deceptive(x) = \begin{cases} u(x) & if\ u(x) > 0 \\ l+1 & if\ u(x) = 0 \end{cases} \quad (5)$$

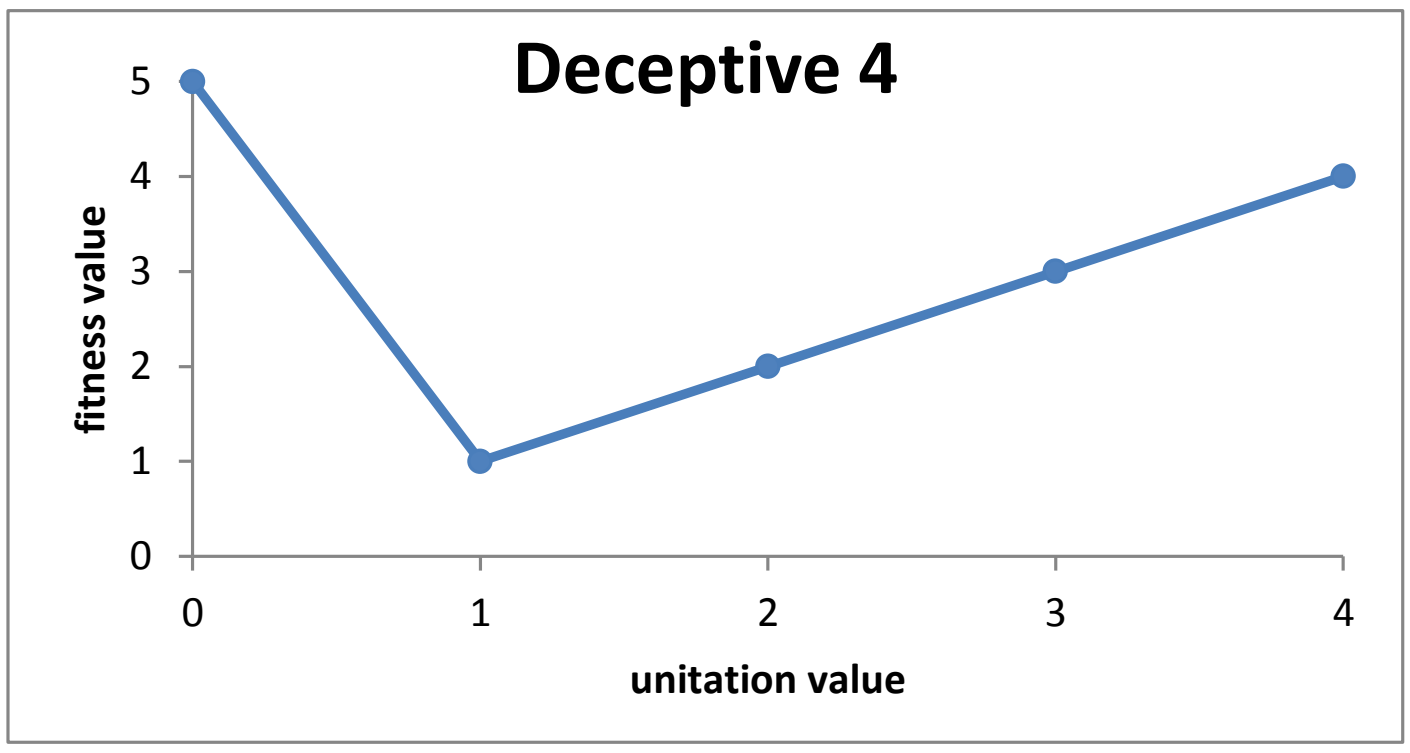

**Fig. 23.** Deceptive function with four bits in the unitation space.

**Table 6**- Statistics of SMuGA and MuGA results in Deceptive 4 function with 16 copies.

| Deceptive 16 4 | SMuGA Mean | SMuGA Std | MuGA Mean | MuGA Std |
|---|---|---|---|---|
| Evals. to Find Best | 5431,04 | 3724,07 | 84749,20 | 30440,50 |
| Best value found | 80,00 | 0,00 | 78,34 | 1,43 |
| Sucess rate (%) | 100,00 | 0,00 | 27,34 | 44,57 |

In this experiment we use a concatenated 16 blocks of four bits deceptive function, **Fig. 23**, representing a chromosome with 64 bits. **Table 6** shows the results of MuGA and SMuGA in the optimization of the function after 100,000 evaluation function calls. SMuGA optimizes all the experiments with very little evaluation function calls when compared to MuGA. **Fig. 24** shows the evolution of the success rate of both algorithms in evolution. MuGA experiences several difficulties in optimizing deceptive functions with large genomes.

Comparing the results with SCA presented in (Wallin et al., 2005), where SCA needs hundreds of thousands of function evaluations, we conclude that SMuGA is significantly better. The ability of SMuGA to manipulate the size of the parasite genomes is the key to solve this kind of problems. SCA do not have that property, and the static size of the parasites slows down the evolution.

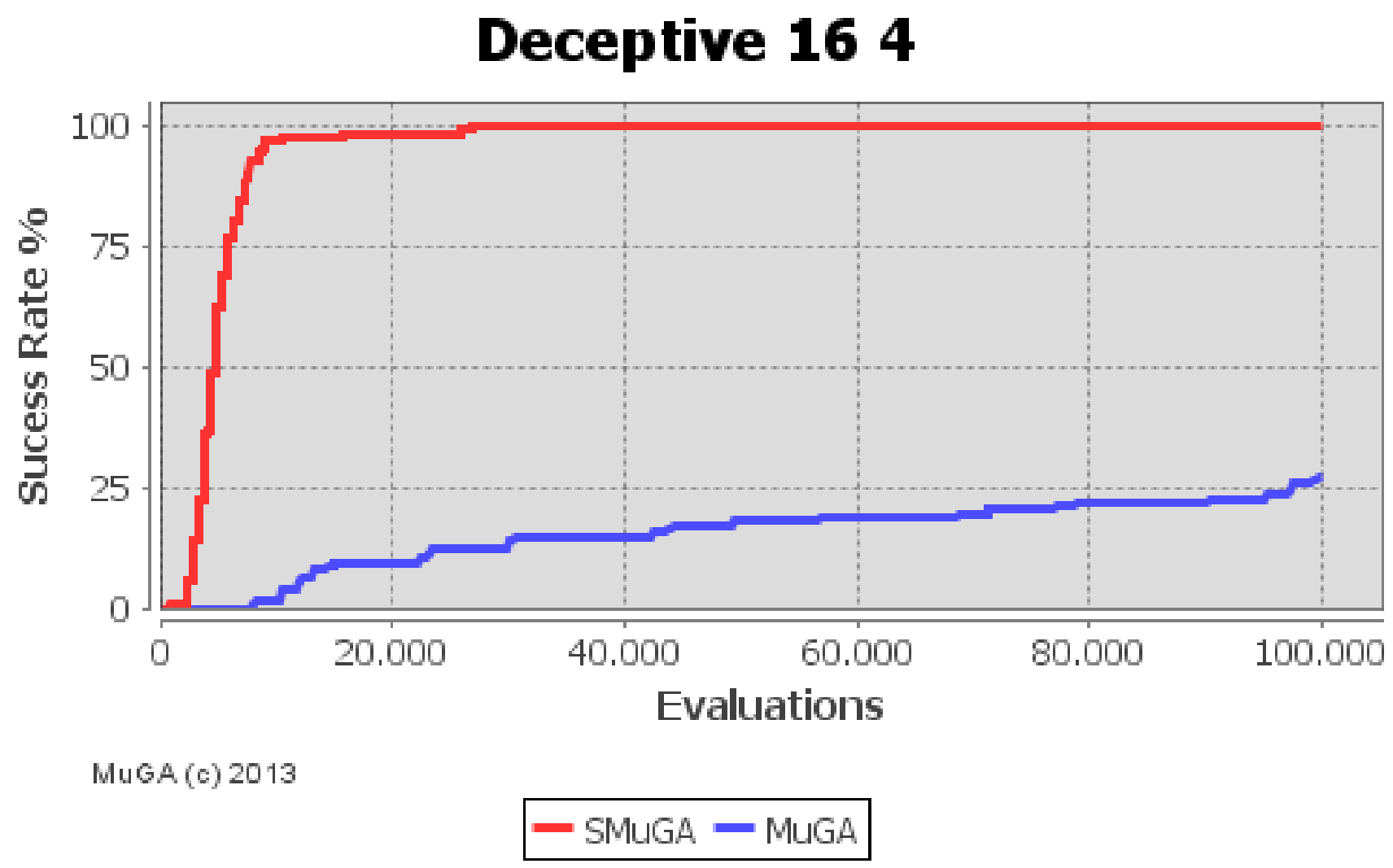

**Fig. 24.** Evolution of the success rate in the optimization of 16 copies of Deceptive 4 function

**Table 7**- Statistics of SMuGA evolution result in optimization of deceptive 4 function with different lengths.

| **SMuGA** | Evals. to find best | | Sucess % | |
|---|---|---|---|---|
| **Deceptive 4** | Mean | Std | Mean | Std |
| 64 bits | 4243.00 | 1645.40 | 100.00 | 0.00 |
| 128 bits | 7061.89 | 3215.18 | 100.00 | 0.00 |
| 256 bits | 11180.36 | 4897.55 | 100.00 | 0.00 |
| 512 bits | 21458.61 | 13696.15 | 96.88 | 17.40 |

**Table 7** and **Fig. 25** show the statistics of evolution after 75,000 function evaluation calls for the problems composed by 16, 32, 64 and 128 deceptive 4 functions that represent genomes with 64, 128, 256 and 512 bits. SMuGA was successful in all the simulations. However in a simulation with problems composed by 512 bits, SMuGA experiments some difficulties in the optimization due to the large genome of the host and more generations are needed to optimize all the problems as we can see in **Fig. 26**.

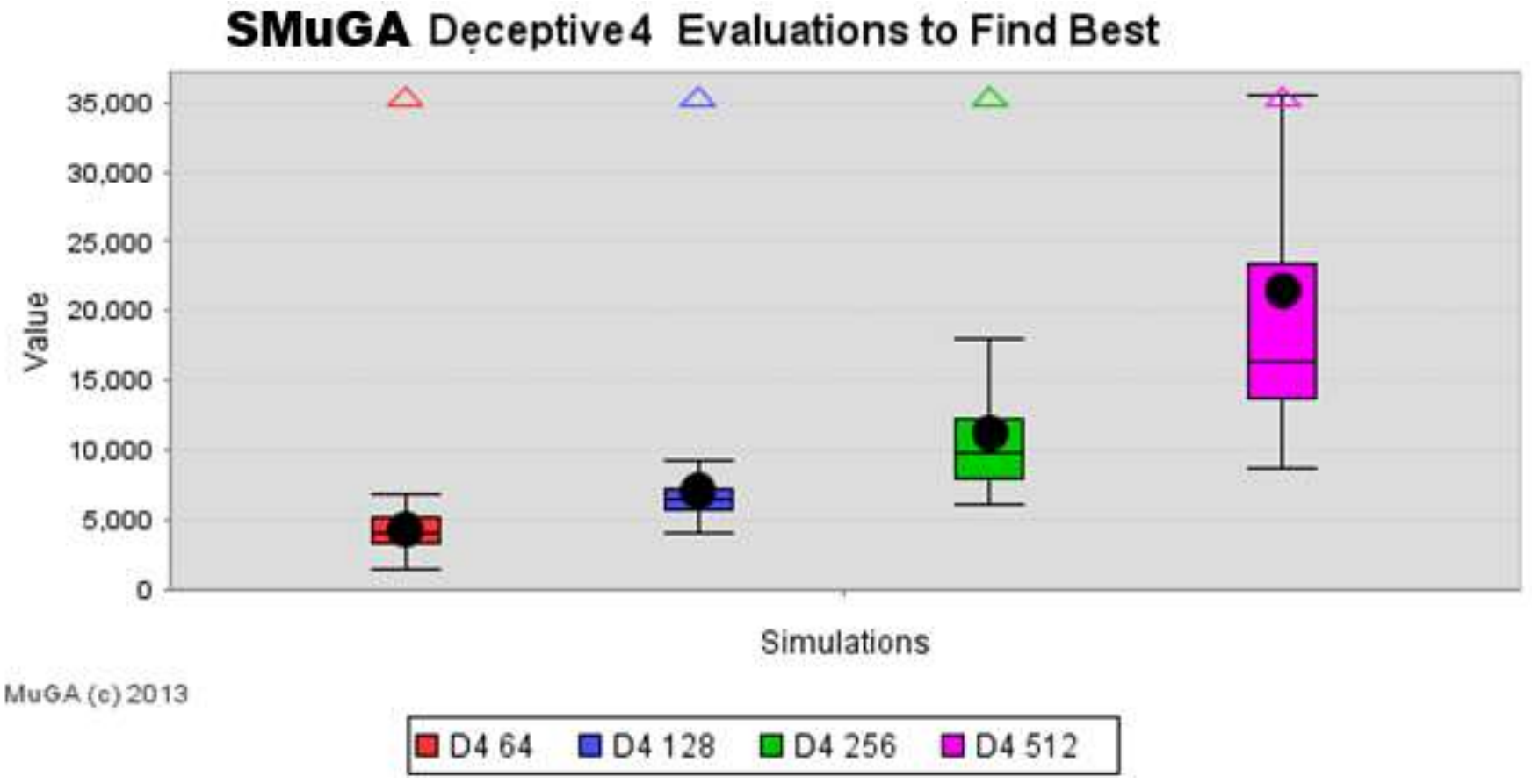

**Fig. 25.** SMuGA: Box-plots of the number of evaluation function calls for SMuGA to find the best value in 16, 32, 64 and 128 copies of deceptive 4 function.

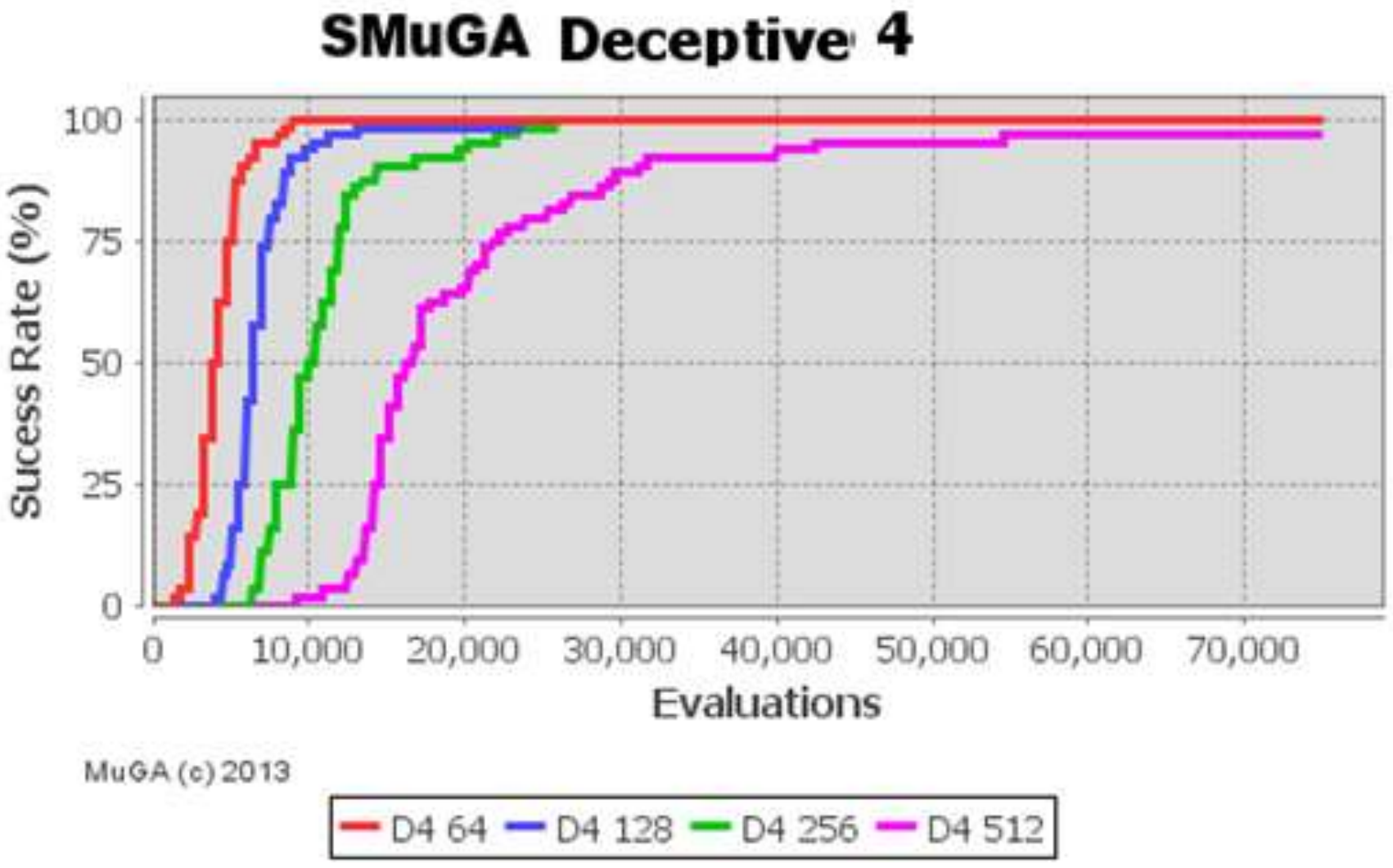

**Fig. 26.** SMuGA: Evolution of the success rate in the optimization of 16, 32, 64 and 128 copies of deceptive 4 function.

SMuGA scales up very well to optimize large deceptive 4 problems and results are again over one order of magnitude better than those presented in (Wallin et al., 2005) using SCA and (Thierens, 2010) using Linkage Tree Genetic Algorithm (LTGA). **Table 8** shows the number of function evaluations to solve Deceptive 4 function with different lengths provided by our best effort to read the graphics supplied in the papers.

**Table 8**- Number of functions evaluation calls to to solve Deceptive4 function using SCA and LTGA algorithm (aprox.).

| Algorithm | Size | Evals. |
|---|---|---|
| SCA | 64 | 100000 |
| | 128 | 200000 |

| Algorithm | Size | Evals. |
|---|---|---|
| LTGA | 60 | 40000 |
| | 100 | 75000 |

### Intertwined Deceptive functions

The Pair-Intertwined function proposed by Wallin and colleagues (Wallin et al., 2005) is defined as two deceptive functions where the bits are intertwined in the same function, **Fig. 27**. The Pair-Intertwined function was many local optima introduced by the combination of the pair of deceptive functions. In this experiment, we use as building block two deceptive functions of four bits each composing a deceptive intertwine function, ***D4PI***, with eight bits.

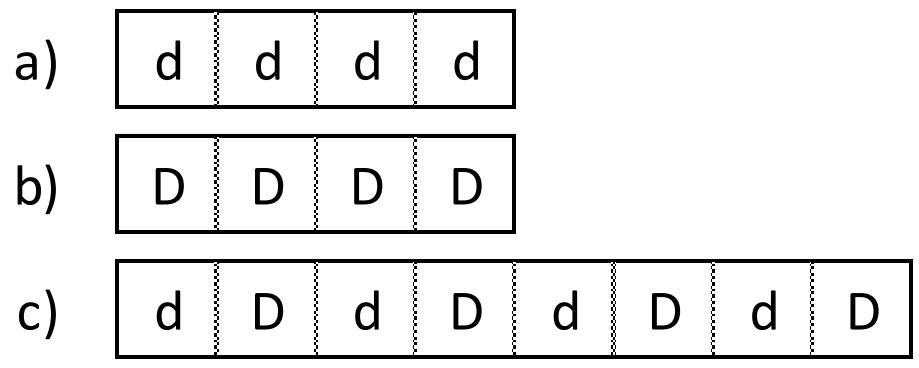

**Fig. 27.** Intertwined pair deceptive functions: a) deceptive function d; b) deceptive function D; c) Intertwined deceptive function dD.

**Table 9**- Statistics of SMuGA and MuGA result in deceptive 4 pair intertwined function with 8 copies

| | SMuGA | | MuGA | |
|---|---|---|---|---|
| D4PI 8 | Mean | Std | Mean | Std |
| Evals. to Find Best | 12401.04 | 12114.49 | 97294.48 | 12106.18 |
| Best value found | 80.00 | 0.00 | 76.88 | 1.61 |
| Sucess rate (%) | 100.00 | 0.00 | 6.25 | 24.21 |

**Table 9** presents the statistics of the optimization of 8 D4PI functions, amounting to 64 bits, after 100,000 evaluation function calls. SMuGA optimizes all the problems with a small number of evaluation function calls due to the capability, provided by the parasites, to discover the BB of the ***D4TI*** function and the ability to concatenate BB and move them along the chromosome. The success of MuGA in this experiment is very limited due to the large length of the BB and the long genome of the individuals, **Fig. 28**. Comparing results with SCA presented in (Wallin et al., 2005), **Table 10**, SMuGA is more than one order of magnitude better.

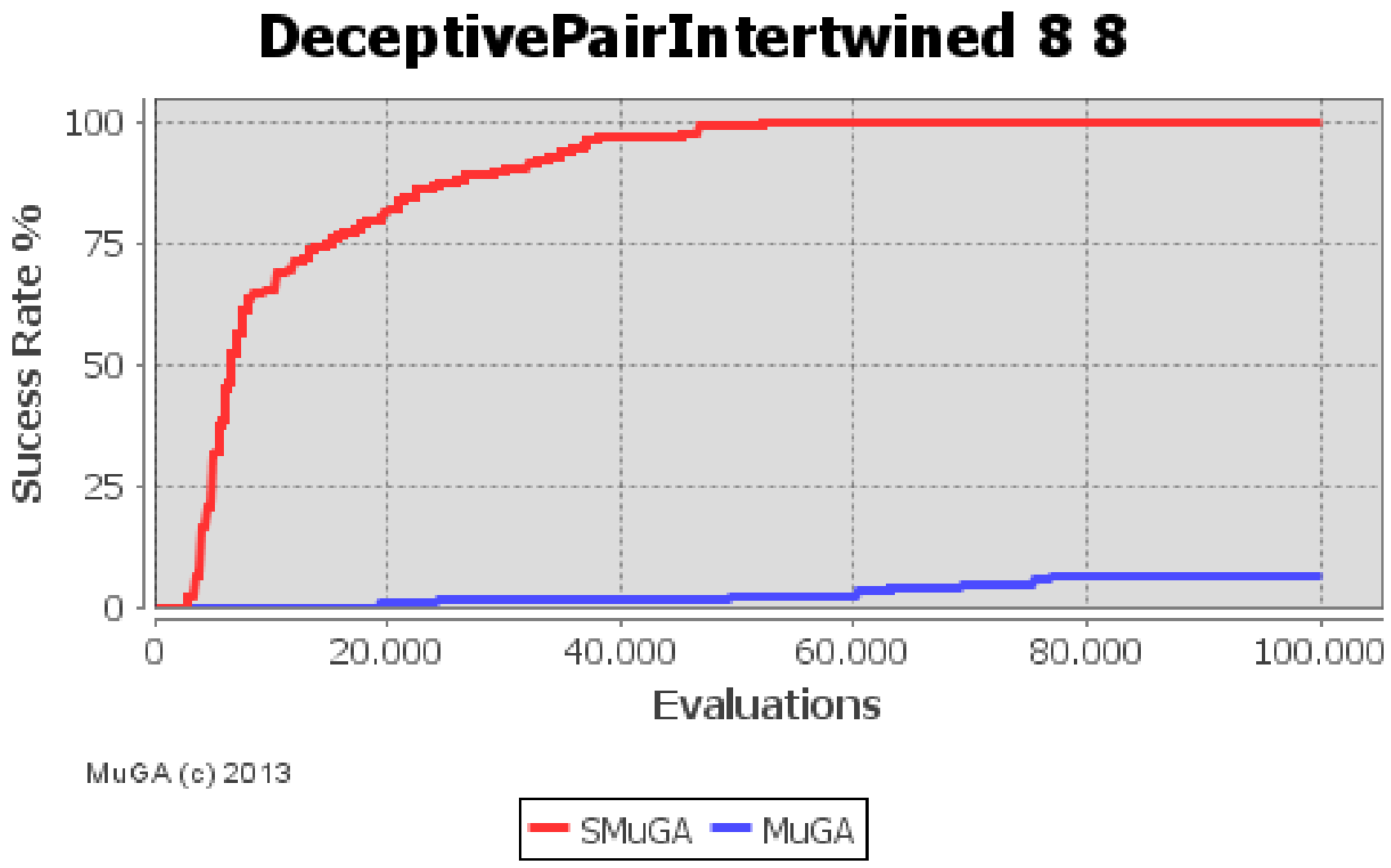

**Fig. 28.** Evolution of the success rate in the optimization of 8 copies of D4PI function

**Table 10**- Number of functions evaluation calls to solve Deceptive Pair Intertwined function using SCA algorithm (aprox.).

| Algorithm | Size | Evals. |
|---|---|---|
| SCA | 64 | 150000 |
| | 128 | 250000 |

**Table 11**- Statistics of SMuGA evolution result in optimization of D4PI with different lengths.

| SMuGA | Evals. to Find best | | Sucess % | |
|---|---|---|---|---|
| D4PI | Mean | Std | Mean | Std |
| 64 bits | 9246.20 | 9045.93 | 100.00 | 0.00 |
| 128 bits | 20050.59 | 21544.04 | 100.00 | 0.00 |
| 256 bits | 56673.89 | 84180.82 | 100.00 | 0.00 |
| 512 bits | 204222.19 | 206614.44 | 79.69 | 40.23 |

**Fig. 29** and **Table 11** show the statistics of evolution after 500,000 function evaluation calls for the problems composed by 8, 16, 32 and 64 D4PI functions which represent chromosomes with 64, 128, 256 and 512 bit. SMuGA was successful in all the simulations. However, in the 512 bit problems SMuGA experiments some difficulties in the optimization due the large genome of the host. More generations would allow to optimize these problems as we can infer from **Fig. 30**, but adjusting the parameters for the 512 bit problem would supposedly increase convergence.

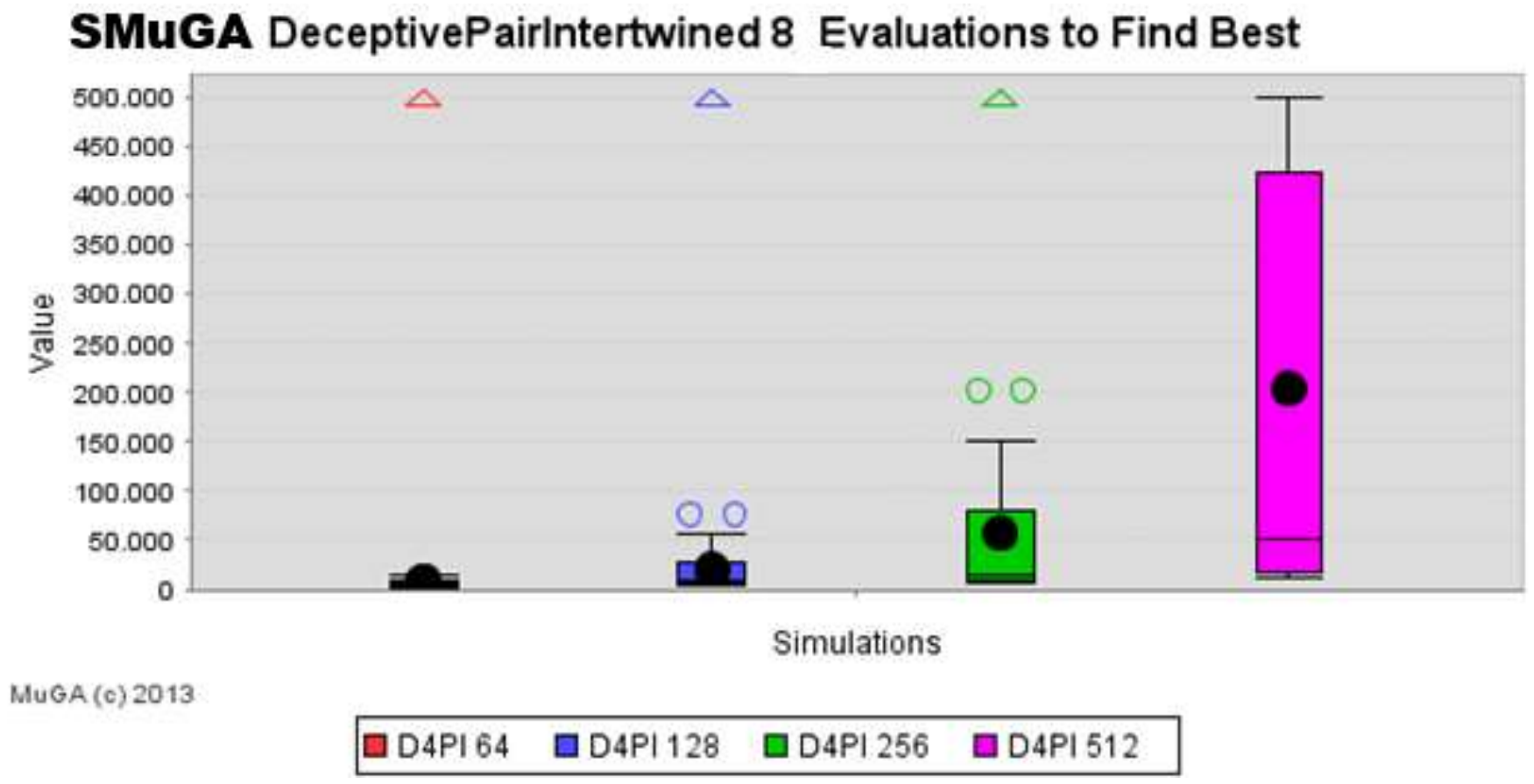

**Fig. 29.** SMuGA: Box-plots of the evaluation function calls to find the best value in 8, 16, 32 and 64 copies of D4PI function.

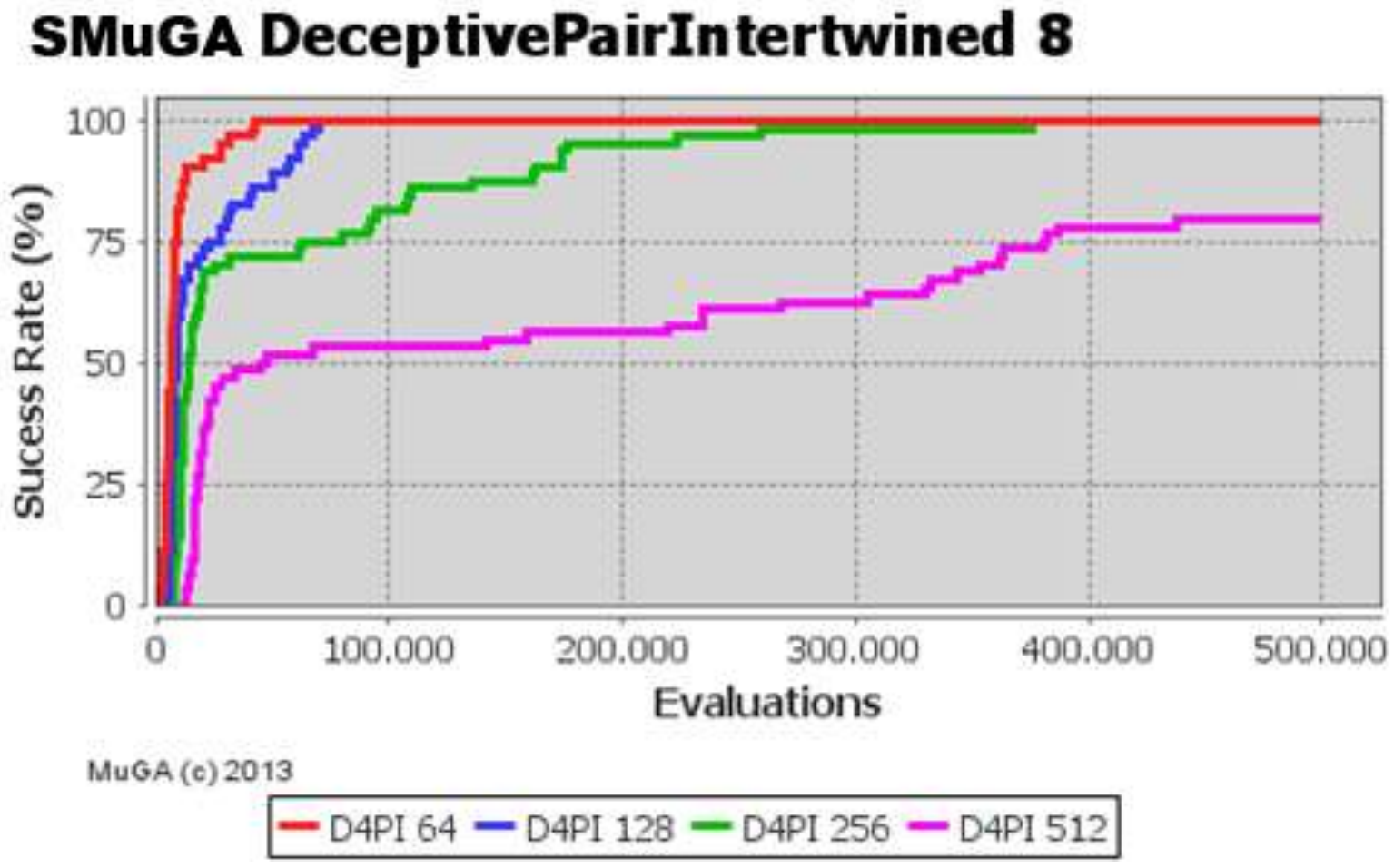

**Fig. 30.** SMuGA: Evolution of the success rate in the optimization of 8, 16, 32 and 64 copies of D4PI function

**Deceptive Intertwined Pair 0-1 function**

To assess the ability of SMuGA to evolve building blocks with optima that are not all ones or all zeroes we defined a new intertwined function, DeceptivePI01, where one function is evaluated by equation (5) and other by the equation (6). In the ***deceptiveZ*** function, equation (6), ***z(x)*** counts the number of zeroes in the string ***x***. The optimum of function DeceptivePI01 is composed by a string with alternating zeroes and ones and the translocations of the building blocks done by the parasites need alignment in the host.

$$deceptiveZ(x) = \begin{cases} z(x) & if\ z(x) > 0 \\ l+1 & if\ z(x) = l \end{cases} \quad (6)$$

**Fig. 25** show the evolution of success rate along the 1,000,000 function evaluation calls for the problems composed by 16, 32, 64 and 128 DeceptivePI01 functions that represent genomes with 64, 128, 256 and 512 bits.

Using parameters of **Table 1** MuGA again shows a poor performance. SMuGA in most simulations optimizes the DeceptivePI01 function composed by eight bit blocks, four of equation 5 and four of equation 6 interleaved. One reason for the failures could be explained by the small number of parasites in the parasite population.

The need for parasite alignment with the host requests a larger population of parasites to avoid local maxima introduced by the bit pattern of the DeceptivePI0 functions. The two local maxima, all ones and all zeroes, are more attractive to the parasites because that pattern does not need alignment and that parasites are easily assimilated by the hosts.

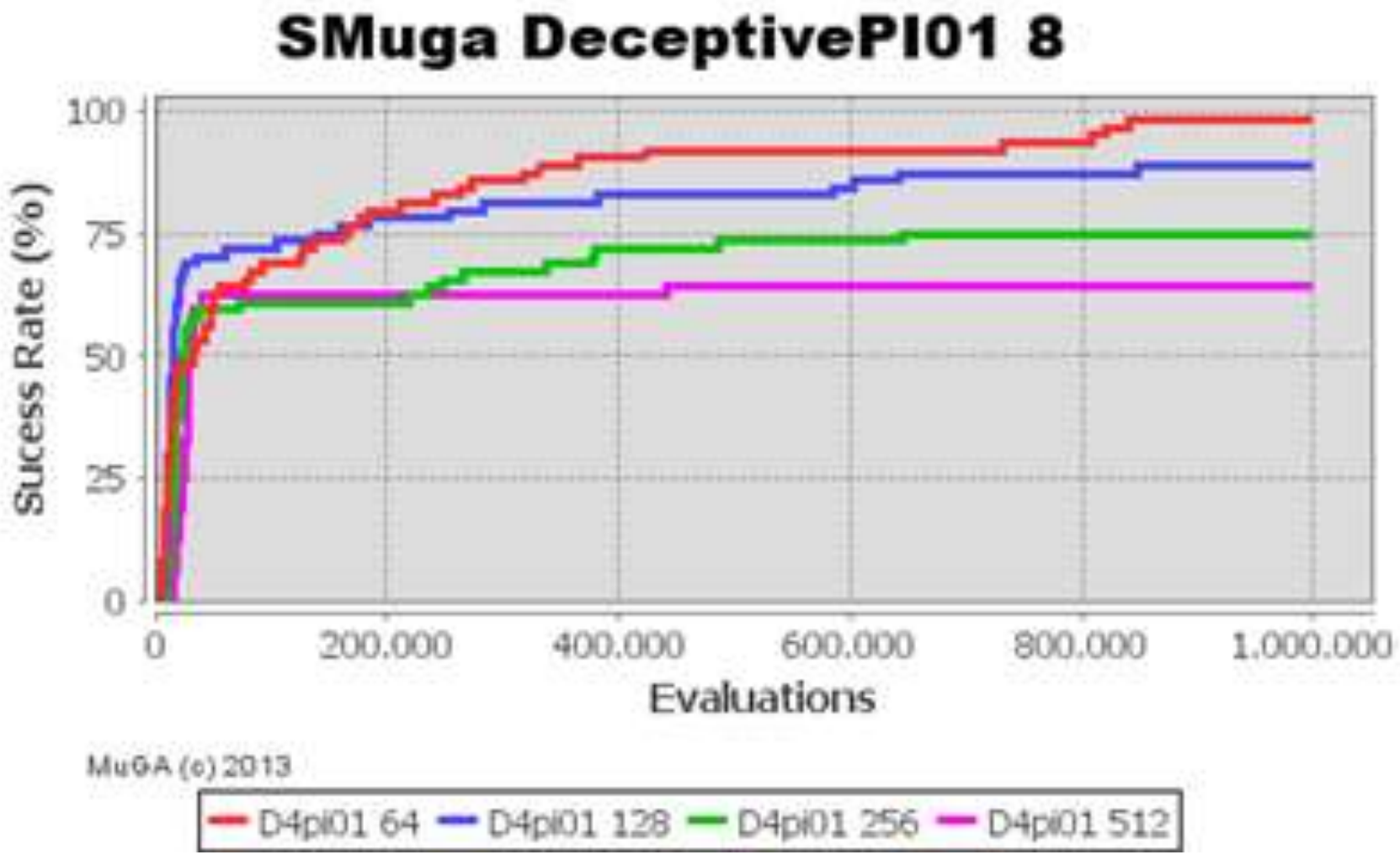

**Fig. 31.** SMuGA: Evolution of the success rate in the optimization of 8, 16, 32 and 64 copies of DeceptivePI01 function with 8 bits.

**Fig. 32** shows the effect of the size of parasite population in the optimization of 8 copies of DeceptivePI01 with 8 bits. As can be seen, the increase of the number of parasites in the symbiotic system increases the robustness of the solver. The increase of parasite population increases the computational complexity of the algorithm, but parasite population can evolve in parallel to the host population exploring the multicore resources of the computers.

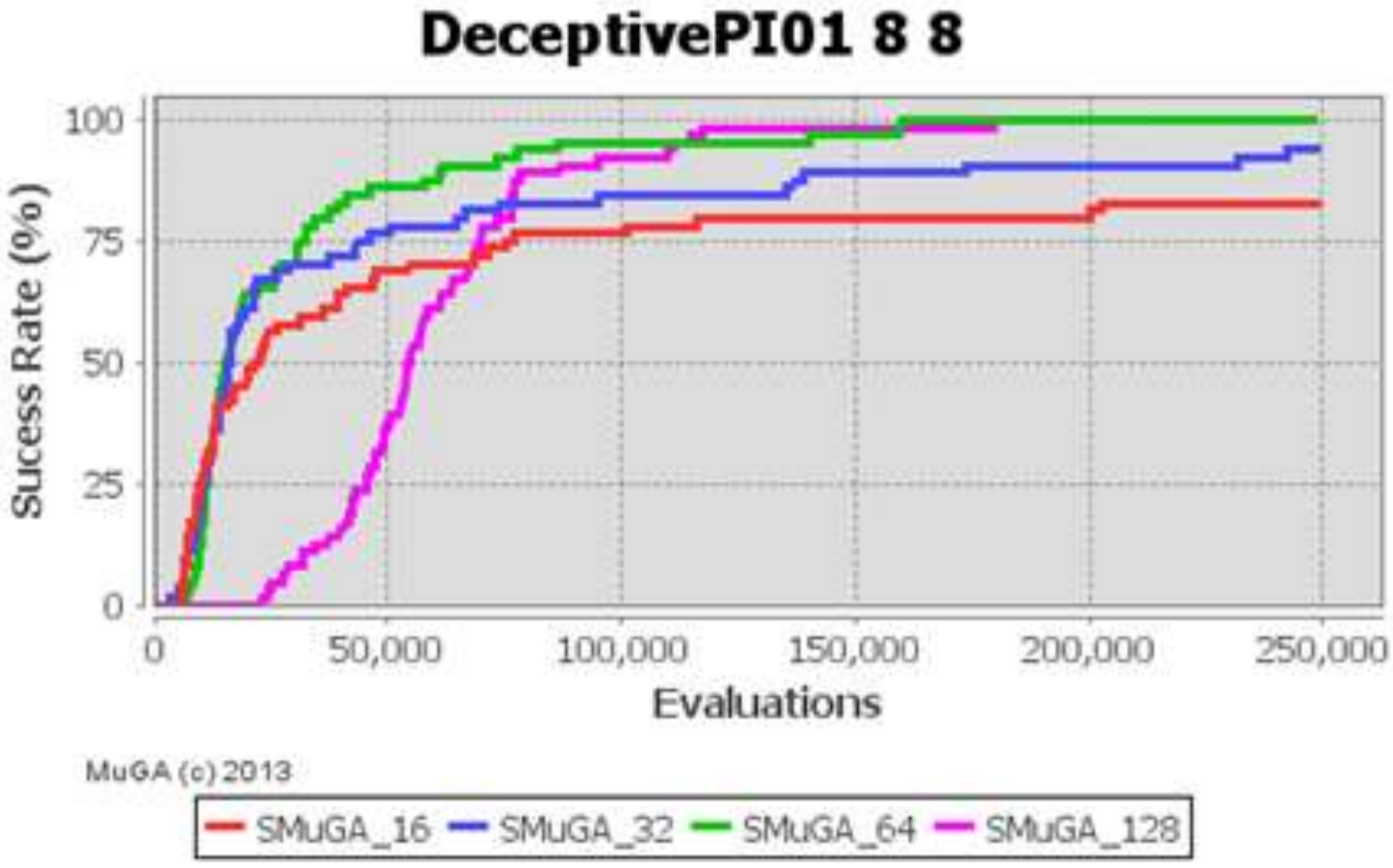

**Fig. 32.** SMuGA: Evolution of the success rate in the optimization of 8 copies of DeceptivePI01 function with 8 bits with solver with 16, 32, 64 and 128 parasites in the parasite population.

**Table 12**- Statistics of SMuGA evolution result in optimization of D4PI01 with different lengths.

| **SMuGA** | Evals. to Find best | | Sucess % | |
|---|---|---|---|---|
| **D4PI01** | Mean | Std | Mean | Std |
| 64 bits | 61204.72 | 29939.75 | 100.00 | 0.00 |
| 128 bits | 54285.36 | 111617.89 | 100.00 | 0.00 |
| 256 bits | 107131.61 | 187231.57 | 100.00 | 0.00 |
| 512 bits | 350228.05 | 421266.94 | 73.44 | 44.17 |

**Fig. 33** and **Table 12** present the same situation of **Fig. 31** but now with 128 elements in the parasite population, instead of 32. The success of the algorithm is increased and simulations evolving functions with 64, 128 and 256 bit are always successfully optimized. The rate of success of simulation with 512 bits also increases although not attaining 100% success. Further parameter tuning is one possible solution to achieve perfect score.

These results show that a large size of the parasite population makes SMuGA more robust in the evolution of difficult functions. Complex bit patterns impose difficulties to SMuGA in the alignment of parasites but these seem to be circumvented by larger parasite populations.

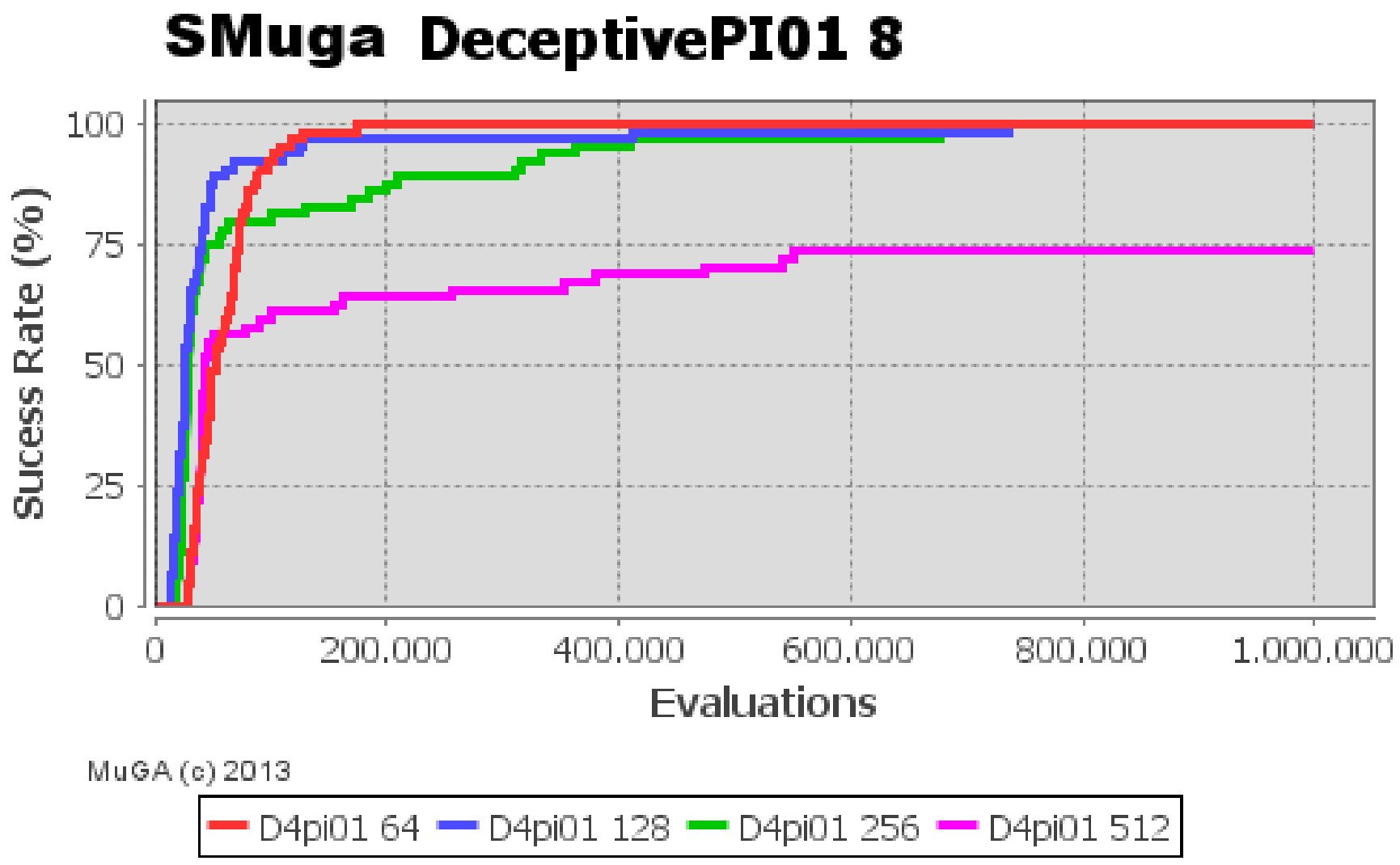

**Fig. 33.** SMuGA: Evolution of the success rate in the optimization of 8, 16, 32 and 64 copies of DeceptivePI01 function with 8 bits intertwined with 128 parasites solvers.

## Conclusions

This chapter presented the Symbiogenetic MuGA (SMuGA), an extension of the Multiset Genetic Algorithm (MuGA) with a novel approach to artificial symbiogenesis where a host receives genetic material from multiple parasites of variable length. This is the first evolutionary model where parasites do not have a fixed length. Rather their length varies along the evolutionary process.

The model proposed also introduced a two-phased step of evolution. In one phase, symbiotic collaborations are generated and compete with previous hosts to form the next generation host population. In the other phase, host and parasite populations evolve on their own for a few generations, but parasites use hosts' fitness as proxys to compute their own. Proxy parasite evaluation significantly saves fitness function calls and avoids the need to generate an exponential number of collaborations. The phase of separate evolution of both hosts and parasites allows to simultaneously stabilize host population and to foster exploration by the parasite population.

Results obtained have largely surpassed previous symbiogenetic models, allowing us to solve very large deceptive problems. It should be noted in spite of MuGA obtaining good results, it is only SMuGA that achieves solutions to very large problems, by integrating symbiogenesis in MuGA, with two-phase evolution and proxy evaluation of parasites.

In fact, SMuGA turned out to be so efficient as to show a linear scaling with the length of the deceptive problems used for testing. The variation of the parasites' length allows evolution to find adequate length building blocks (BB) for the problem at hand. Accumulating multiple parasites in a single host provides the opportunity of using parasite combinations, which prove to be important for more complex problems.

In future work, we want to test more operators in the parasites. In particular, inversion might be important to hierarchical deceptive problems. We also need to explore different types of problems with SMuGA. Those used in this paper are repeated concatenations of the same function. Also, the flexibility of this model indicates that it is adequate for dynamic fitness functions, and we should test it on dynamic problems. The symbiotic system can also be taken as a new operator introducing new parameters in the evolutionary process. Consequently the new parameters can be tuned to increase the effectiveness of SMuGA and in the future we will make an effort in optimization and automation of these parameters.

## Acknowledgments

The authors thank Mel Todd, Guida Manso and Nathalie Gontier for the precious revisions that made this text more clear and readable.

## References

Abdoun, O., Abouchabaka, J., Tajani, C., 2012. Analyzing the Performance of Mutation Operators to Solve the Travelling Salesman Problem. arXiv:1203.3099.

Ackley, D.H., 1987. A Connectionist Machine for Genetic Hillclimbing. Kluwer Academic Publishers, Norwell, MA, USA.

Baghshah, M.S., Shouraki, S.B., Halavati, R., Lucas, C., 2007. Evolving fuzzy classifiers using a symbiotic approach, in: IEEE Congress on Evolutionary Computation, 2007. CEC 2007. Presented at the IEEE Congress on Evolutionary Computation, 2007. CEC 2007, pp. 1601–1607.

Chen, Y., Hu, J., Hirasawa, K., Yu, S., 2008. Solving deceptive problems using a genetic algorithm with reserve selection, in: IEEE Congress on Evolutionary Computation, 2008. CEC 2008. (IEEE World Congress on Computational Intelligence). Presented at the IEEE Congress on Evolutionary Computation, 2008. CEC 2008. (IEEE World Congress on Computational Intelligence), pp. 884 –889.

Daida, J.M., Grasso, C.S., Stanhope, S.A., Ross, S.J., 1996. Symbionticism and Complex Adaptive Systems I: Implications of Having Symbiosis Occur in Nature, in: Proceedings of the Fifth Annual Conference on Evolutionary Programming. The MIT Press, pp. 177–186.

Droste, S., Jansen, T., Wegener, I., 2002. On the analysis of the (1+ 1) evolutionary algorithm. Theor. Comput. Sci. 276, 51–81.

Dumeur, R., 1996. Evolution through cooperation: The Symbiotic Algorithm, in: Alliot, J.-M., Lutton, E., Ronald, E., Schoenauer, M., Snyers, D. (Eds.), Artificial Evolution, Lecture Notes in Computer Science. Springer Berlin Heidelberg, pp. 145–158.

Goldberg, D.E., 1987. Simple Genetic Algorithms and the Minimal, Deceptive Problem. Genet. Algorithms Simulated Annealing L. Davis, editor, San Mateo, CA: Morgan Kaufmann, 74–88.

Goldberg, D.E., 1989. Genetic Algorithms and Walsh Functions: Part I, A Gentle Introduction. Complex Syst. 3, 129–152.

Herrera, F., Lozano, M., Sanchez, A.M., 2003. A taxonomy for the crossover operator for real-coded genetic algorithms: An experimental study. Int. J. Intell. Syst. 18, 309–338.

Heywood, M.I., Lichodzijewski, P., 2010. Symbiogenesis as a Mechanism for Building Complex Adaptive Systems: A Review, in: Chio, C.D., Cagnoni, S., Cotta, C., Ebner, M., Ekárt, A., Esparcia-Alcazar, A.I., Goh, C.-K., Merelo, J.J., Neri, F., Preuß, M., Togelius, J., Yannakakis, G.N. (Eds.), Applications of Evolutionary Computation, Lecture Notes in Computer Science. Springer Berlin Heidelberg, pp. 51–60.

Jayachandran, J., Corns, S., 2010. A comparative study of diversity in evolutionary algorithms. IEEE, pp. 1–7.

Lozano, M., Herrera, F., Cano, J., 2008. Replacement strategies to preserve useful diversity in steady-state genetic algorithms. Inf. Sci. 178, 4421–4433.

Manso, A., Correia, L., 2009. Genetic algorithms using populations based on multisets, in: New Trends in Artificial Intelligence, EPIA 2009. Luís Seabra Lopes and Nuno Lau and Pedro Mariano and Luís Rocha (Eds.), Universidade de Aveiro, pp. 53–64.

Manso, A., Correia, L., 2011. A multiset genetic algorithm for real coded problems, in: Proceedings of the 13th Annual Conference Companion on Genetic and Evolutionary Computation - GECCO '11. Presented at the the 13th annual conference companion, Dublin, Ireland, p. 153.

Manso, A., Correia, L., 2013. A multiset genetic algorithm for the optimization of deceptive problems, in: Proceeding of the Fifteenth Annual Conference on Genetic and Evolutionary Computation Conference, GECCO ’13. ACM, New York, NY, USA, pp. 813–820.

Otman, A., Jaafar, A., 2011. A Comparative Study of Adaptive Crossover Operators for Genetic Algorithms to Resolve the Traveling Salesman Problem. IJCA (0975-8887).

Paredis, J., 1995. The Symbiotic Evolution of Solutions and Their Representations, in: Proceedings of the 6th International Conference on Genetic Algorithms. Morgan Kaufmann Publishers Inc., San Francisco, CA, USA, pp. 359–365.

Rosin, C.D., Belew, R.K., 1997. New Methods for Competitive Coevolution. Evol Comput 5, 1–29.

Sivaraj, R., Ravichandran, T., 2011. A review of selection methods in genetic algorithm. Int. J. Eng. Sci. Technol. 3, 3792–3797.

Spears, W., Anand, V., 1991. A Study Of Crossover Operators In Genetic Programming.

Thierens, D., 2010. Linkage Tree Genetic Algorithm: First Results, in: Proceedings of the 12th Annual Conference Companion on Genetic and Evolutionary Computation, GECCO ’10. ACM, New York, NY, USA, pp. 1953–1958.

Wallin, D., Ryan, C., Azad, R.M.A., 2005. Symbiogenetic coevolution, in: The 2005 IEEE Congress on Evolutionary Computation, 2005. Presented at the The 2005 IEEE Congress on Evolutionary Computation, 2005, pp. 1613–1620 Vol. 2.

Whitley, L.D., 1991. Fundamental Principles of Deception in Genetic Search. Found. Genet. Algorithms 1, 221–241.

Yang, S., 2004. Adaptive group mutation for tackling deception in genetic search. WSEAS Trans. Syst. 3, 107–112.

Yu, E.L., Suganthan, P.N., 2010. Ensemble of niching algorithms. Inf. Sci. 180, 2815–2833.